\newif\ifsingle
\singletrue 

\newif\ifFullVersion
\FullVersiontrue 

\ifsingle
\documentclass[12pt,draftclsnofoot, onecolumn]{IEEEtran}		
\else		
\documentclass[10pt,final, twocolumn]{IEEEtran}
\fi

\usepackage{times}
\usepackage{amsmath,dsfont}
\usepackage{amssymb,amsthm}
\usepackage{epsfig,verbatim}
\usepackage{caption}
\usepackage{subcaption}
\usepackage{setspace}
\usepackage{color}
\usepackage{cite}
\usepackage{epstopdf}
\usepackage{graphicx}
\usepackage{accents}
\usepackage{acronym}
\usepackage[bookmarks,colorlinks]{hyperref}
\usepackage{booktabs}
\usepackage{mathtools} 
\usepackage{enumitem}

\usepackage[ruled,linesnumbered]{algorithm2e}  
\SetKwInput{KwData}{\textbf{Input}} 

\newcommand{\myVec}[1]{{\boldsymbol{#1}}}

\newcommand{\mySet}[1]{\mathcal{#1}}
\newcommand{\myX}{{\myVec{x}}}			 		
\newcommand{\myZ}{{\myVec{z}}}	
\newcommand{\myY}{\myVec{y}}
\newcommand{\Nusers}{K}

\newcommand{\capacity}{C}
\newcommand{\Nparameters}{M}
\newcommand{\NLabels}{N}
\newcommand{\MyWeights}{\myVec{\theta}}
\newcommand{\CommLink}{L}

\newcommand{\NetMap}{\myVec{f}}
\newcommand{\bits}{B}
\newcommand{\NusersSet}{\mySet{K}}			 			

\newcommand{\norm}[1]{\left\lVert#1\right\rVert}

\newtheorem{theorem}{Theorem}
\newtheorem{corollary}{Corollary}

\ifsingle
\newcommand{\figWidth}{0.8\columnwidth}

\else
\newcommand{\figWidth}{\columnwidth}

\fi 

\acrodef{adc}[ADC]{analog-to-digital convertor}
\acrodef{cs}[CS]{compressed sensing}
\acrodef{dtft}[DTFT]{discrete-time Fourier transform}
\acrodef{dnn}[DNN]{deep neural network} 
\acrodef{csi}[CSI]{channel state information}
\acrodef{map}[MAP]{maximum a-posteriori probability}
\acrodef{snr}[SNR]{signal-to-noise ratio}
\acrodef{bs}[BS]{base station} 
\acrodef{iot}[IOT]{Interent of Things}
\acrodef{mimo}[MIMO]{multiple-input multiple-output}
\acrodef{mse}[MSE]{mean-squared error}
\acrodef{pdf}[PDF]{probability density function}
\acrodef{rv}[RV]{random variable}
\acrodef{fec}[FEC]{forward error correction}
\acrodef{dma}[DMA]{dynamic metasurface antenna}
\acrodef{lti}[LTI]{linear time-invariant}
\acrodef{wss}[WSS]{wide-sense stationary}
\acrodef{psd}[PSD]{power spectral density}
\acrodef{ser}[SER]{symbol error rate} 
\acrodef{ber}[BER]{bit error rate} 
\acrodef{sgd}[SGD]{stochastic gradient descent} 
\acrodef{isi}[ISI]{intersymbol interference}  
\acrodef{awgn}[AWGN]{additive white Gaussian noise} 
\acrodef{ut}[UT]{user terminal} 
\acrodef{mmw}[mmWave]{millimeter wave}
\acrodef{ai}[AI]{artifical intelligence}
\acrodef{cb}[$\mySet{Q}$]{codebook}
\acrodef{vqvae}[VQ-VAE]{vector quantized variational autoencoder}

\IEEEoverridecommandlockouts

\title{Decentralized Low-Latency Collaborative Inference via Ensembles on the Edge
}
\author{  
	\IEEEauthorblockN{May Malka, Erez Farhan, Hai Morgenstern, and Nir Shlezinger\\
	} 
	\thanks{
		Parts of this work were presented in the IEEE International Conference on Acoustics, Speech, and Signal Processing (ICASSP) 2021 as the paper \cite{shlezinger2021collaborative}.
		This work was supported in part by the BGU data science center. 
		M. Malka and N. Shlezinger are with the School of ECE, Ben-Gurion University of the Negev, Beer-Sheva, Israel (e-mail: maymal@post.bgu.ac.il; nirshl@bgu.ac.il).
		E. Farhan is with Facebook (e-mail: erezfarhan@gmail.com).
		H. Morgenstern is with  Ramon.Space, Israel (e-mail: hai.morgenstern@ramon.space). }

	\vspace{-1.0cm}
	
}
\vspace{-0.75cm}

\begin{document}
	
	\maketitle
	\pagestyle{plain}
	\thispagestyle{plain}
	\begin{abstract} 
        The success of deep neural networks (DNNs) is heavily dependent on computational resources. While DNNs are often employed on cloud servers, there is a growing need to operate DNNs on edge devices. Edge devices are typically limited in their computational resources, yet, often multiple edge devices are deployed in the same environment and can reliably communicate with each other.  In this work we propose to facilitate the application of DNNs on the edge by allowing multiple users to collaborate during inference to improve their accuracy. Our mechanism, coined {\em edge ensembles}, is based on having diverse predictors at each device, which form an ensemble of models during inference. To mitigate the communication overhead,  the users share quantized features, and we propose a method for aggregating multiple decisions into a single inference rule. We analyze the latency induced by edge ensembles, showing that its performance improvement comes at the cost of a minor additional delay under common assumptions on the communication network. Our experiments demonstrate that collaborative inference via edge ensembles equipped with compact DNNs substantially improves the accuracy over having each user infer locally, and can outperform using a single centralized DNN larger than all the networks in the ensemble together.
		
	\end{abstract}
	
	\vspace{-0.4cm}
	\section{Introduction}
	\vspace{-0.1cm} 
	
	Deep learning systems have demonstrated unprecedented success in various applications, including computer vision, natural language processing, and speech \cite{lecun2015deep}. Such systems employ \acp{dnn} that consist of highly-parameterized models, trained using massive volumes of data.  Consequently, deep learning is traditionally the domain of large scale computer servers, which have the computational resources and the ability to aggregate the data required to store, train, and apply \acp{dnn}.

	While \acp{dnn} are traditionally trained and maintained on powerful computer servers, the interface with the real world is typically carried out using {\em edge devices}. The data used for inference, including images, text messages, and signal recordings, is obtained on the user's side via devices such as smartphones, tablets, sensors, and autonomous vehicles. This motivates the implementation of \acp{dnn} on edge devices \cite{chen2019deep}.  Applying \acp{dnn} as a form of mobile edge computing allows to infer on the same device where the data is collected, rather than having the samples sent to a centralized cloud server. Such \ac{dnn}-aided edge devices can operate at  various connectivity conditions with reduced latency, as well as alleviate privacy issues and facilitate the personalization of  \ac{ai} systems \cite{mao2017survey}. As a result, recent years have witnessed growing interest in  implementing deep learning and deploying \acp{dnn} on edge device \cite{chen2019deep, mao2017survey,zhou2019edge, zhang2019deep, mcmahan2016communication,gafni2021federated, li2019federated}.

	One of the main challenges associated with implementing trained \acp{dnn} on edge devices stems from their limited computational resources \cite{xu2018scaling}. A leading strategy to tackle this challenge is to make \acp{dnn} more compact. This is achieved by compressing existing highly-parameterized \acp{dnn} via pruning \cite{cheng2017survey, he2017channel} and quantization \cite{agustsson2017soft,jacob2018quantization} schemes, or alternatively, by designing \ac{dnn}-aided systems to utilize compact networks by incorporating statistical model-based domain knowledge \cite{shlezinger2020inference,farsad2020data,shlezinger2020model}. Such strategies focus on a single edge user, and thus do not exploit the fact that while each device is limited in its hardware, multiple users can confidently and securely collaborate to benefit from their joint computational resources. 
	
	An alternative approach which accounts for the ability of edge devices to collaborate is based on partitioning and dividing a highly-parameterized \ac{dnn}  among multiple devices. These can be a group of edge devices which can communicate with one another, or the edge user and its server or access point. The partitioning of a multi-layered \ac{dnn} among multiple users to jointly form the large network during inference is typically referred to as {\em computation offloading} \cite{lin2019computation,chen2019deep,mao2017survey}, while the division of a \ac{dnn} between an edge device and its edge server is commonly coined {\em collaborative intelligence} \cite{bajic2021collaborative, cohen2021lightweight}. In both cases, each participating device only maintains a subset of the layers of the \ac{dnn}, and communicates its output features, which are possibly compressed to reduce the overhead \cite{cohen2021lightweight, merluzzi2021dynamic}, to the specific device which maintains the subsequent layers. The main drawback of this approach is that each user cannot infer on its own, and the complete set of devices among which the \ac{dnn} is divided must be present. This results in high dependence on connectivity and possibly increased latency.
	
	The dynamic and mobile nature of edge devices and their ability to communicate with each other in a device-to-device fashion or via mobile ad-hoc networks \cite{burbank2006key} indicates that \ac{dnn}-aided edge devices should be able to both infer locally (for limited connectivity scenario) as well as benefit from collaboration among multiple users. Nonetheless, the  above mentioned existing approaches either rely solely on local inference (via compact \acp{dnn}), or rely on constant reliable communications with a predefined set of edge devices (in computation offloading) or an edge server (for collaborative intelligence). 
	This motivates designing protocols and architectures for \ac{dnn}-aided edge devices that are adaptive to different conditions in the communication network. Such protocols should allow each user to infer on its own while supporting collaboration between a varying number of different neighboring devices when communications is attainable in a purely decentralized manner.

	In this work we propose a framework for utilizing \acp{dnn} on edge devices by leveraging device-to-device communications to allow multiple users to form a {\em deep ensemble}. Deep ensembles are scalable \ac{dnn} architectures comprised of a multitude of diverse deep predictors, which carry out inference by aggregating the individual predictions \cite{sagi2018ensemble}. Deep ensembles were empirically shown to achieve high accuracy and generalization performance in a manner that improves with the number of individual \acp{dnn} \cite{lakshminarayanan2017simple, chen2020group, li2019ensemblenet, brazowski2020collective, raviv2020data, fort2019deep}.  Here, we show that the concept of deep ensembles gives rise to a collaborative inference method for \ac{ai}-empowered edge devices, referred to as {\em edge ensembles}, which allows users to benefit from the presence of communication links with other devices to improve accuracy while being operable in limited connectivity environments. 
	
	We present a decentralized collaborative inference protocol which allows each user to form a deep ensemble with its neighboring devices during inference. To be able to benefit from collaboration, we propose to partition the \ac{dnn} architecture into a shared encoder and user-specific decoder, such that diversity among the models, which is key to successful collaboration, is achieved. To reduce the communication overhead during inference and its associated latency, we employ a trainable quantizer applied to the encoded features, resulting in the users exchanging a limited number of bits during inference stage. Such setups harness the capabilities of each user to infer locally, while allowing the users to operate in the absence of connectivity. Our proposed collaborative inference method can be applied to a variable number of devices,  while benefiting from the presence of neighbouring users with low latency by sharing only quantized features.
	
	We propose two methods for collaborative inference with quantized features. In the first  approach all participating users process the quantized features, thus requiring each device to maintain a single deep model; the second technique requires only the neighbouring devices to process the quantized features and has the aggregating user process the unquantized sample. The latter approach utilizes a dedicated quantization-aware aggregation mechanism, which allows to achieve improved accuracy, while requiring the users to maintain two deep models.
	We analyze the delay associated with edge ensembles, showing that the parallel nature of deep ensembles in which all \acp{dnn} predict simultaneously, results at an improved accuracy with a minor additional delay compared with having single devices inferring on their own.
	
	Our experimental study considers edge devices utilizing the  MobilenetV2 deep classifier \cite{sandler2018mobilenetv2}, applied to images taken from the CIFAR-10, CIFAR-100, and ImageWoof datasets.
	We carry out extensive numerical evaluations which demonstrate that the proposed decentralized low-latency collaborative inference protocol can notably improve the  accuracy compared to inferring locally, while mitigating the dependence of edge devices on reliable communications with a centralized server.
	Furthermore, we show that in some scenarios, edge ensembles, in which each user has a limited amount of weights, can achieve improved accuracy over a single centralized \ac{dnn} with more parameters than those of all the devices together. 
	
	The rest of this paper is organized as follows;  in Section~\ref{sec:Model}, we detail the problem of \ac{dnn}-based edge inference and review some basics in deep ensembles. Section~\ref{sec:EdgeEnsembles} presents the  edge ensembles mechanism and analyzes its latency. Experimental results are stated in Section~\ref{sec:Sims}. Finally, Section~\ref{sec:Conclusions} provides concluding remarks. 	 
	
	Throughout the paper, we use boldface lower-case letters for vectors, e.g., ${\myVec{x}}$;
	the $i$th element of ${\myVec{x}}$ is written as $[{\myVec{x}}]_i$.  
	calligraphic letters, such as $\mySet{X}$, are used for sets, 
	%
	%
	and $\mySet{R}$ is the set of real numbers. 
	The $\ell_2$ norm is denoted by $\| \cdot \|_2$, while $\Pr(\cdot)$ is the probability function. 
	
	\vspace{-0.2cm}
	\section{System Model}
	\label{sec:Model}
	\vspace{-0.1cm}
	In this section we present the system model, under which our proposed edge ensembles protocol is derived in Section~\ref{sec:EdgeEnsembles}. We begin by formulating the problem of \ac{dnn}-based edge inference in Subsection~\ref{subsec:Problem}. Then, we discuss the notion of collaborative inference and its associated requirements in Subsection~\ref{subsec:Inference}. We conclude by reviewing some preliminaries in deep ensembles and in deep vector quantization in Subsections~\ref{subsec:DeepEnsembles} and~\ref{subsec:Quantization}, respectively. 
	
	\vspace{-0.2cm}
	\subsection{Problem Formulation}
	\label{subsec:Problem}
	\vspace{-0.1cm}  
	We consider a set of $\Nusers$ \ac{ai}-empowered mobile edge devices. Each device can communicate with its neighbor devices over dedicated rate-limited communication links. 
	At a given time $t$, user $i_t \in \{1,\ldots,\Nusers\} \triangleq \NusersSet$ receives an input data sample $\myX_{i_t}$, such as an image or an audio recording, and wishes to carry out inference, e.g., classification. The common practice for applying \ac{dnn}-aided inference to data acquired by edge devices requires the user to send the input data to a centralized server that maintains a pre-trained highly parameterized \ac{dnn}. The server carries out inference and conveys its result back to the user. The growing demands for low latency on-device inference in different connectivity conditions, as well as  privacy considerations limiting the sharing of data samples, necessitates the users to be able to infer reliably without relying on a central server.

\begin{figure}
		\centering
		\includegraphics[width = \figWidth]{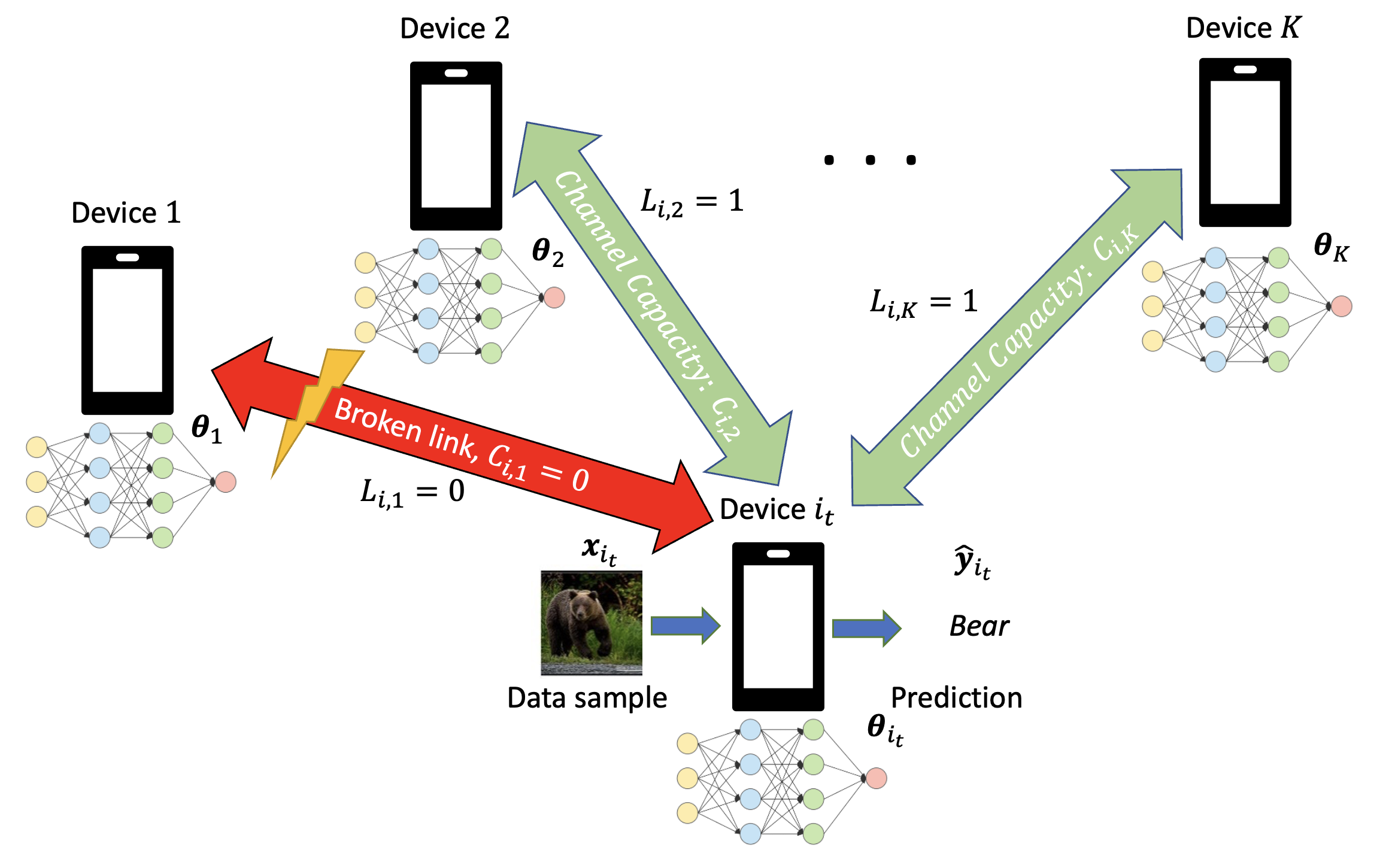}  
		\caption{\ac{dnn}-based edge inference illustration.}
		\label{fig:IllustrationWithCapacity}
	\end{figure}

	In mobile edge computing, the improved computational capabilities of modern edge devices are exploited to allow each user to maintain its local deep model. In particular, the user of index $i \in \NusersSet$ has access to a pre-trained model comprised of $\Nparameters$ parameters, represented by the vector $\MyWeights_i \in \mySet{R}^{\Nparameters}$, such that the model mapping is dented by $f_{\MyWeights_i}(\cdot)$. The users can communicate with each other, e.g., over a wireless communication channel. Due to the dynamic nature of the devices, the communication links may not be stable and have a limited throughput. 
	The quality of the link is represented by its instantaneous capacity, where we use the \ac{rv} $\capacity_{i,j}^t$ for the capacity of the link between the users $i,j\in \NusersSet$ at time instance $t$. If the link between the users $i,j\in \NusersSet$ is unavailable or  broken at time instance $t$, then its capacity is zero. To formulate the links behavior, we define the set of \acp{rv} $\{\CommLink_{i,j}^{t}\}$ representing the status of the communication links between the users $i$ and $j$ at time instance $t$, i.e., $\CommLink_{i,j}^{t} = 0$ implies that $\capacity_{i,j}^t = 0$ while  $\capacity_{i,j}^t > 0$ when $\CommLink_{i,j}^{t} = 1$. We assume reciprocal channels, such that  $\CommLink_{i,j}^{t} = \CommLink_{j,i}^{t}$, and set  $\CommLink_{i,i}^{t}\equiv 1$. Such setups represent, for example, \ac{ai}-empowered vehicles which may operate in areas without connectivity to some centralized cloud server while being able to communicate in a device-to-device manner.  An illustration of the resulting setup is depicted in Fig.~\ref{fig:IllustrationWithCapacity}. 
	
	Our goal is to characterize a mechanism and the corresponding \ac{dnn} architecture for facilitating inference based on the sample $\myX_{i_t}$,  focusing on adaptive schemes which do not require connectivity with a central server.
	We use ${\myY}_{i_t}$ and $\hat{\myY}_{i_t}$ to denote the label value associated with  $\myX_{i_t}$ and its estimate produced by the set of \ac{ai}-empowered mobile devices, respectively, while $\mySet{L}_{e}(\cdot ; \cdot)$ is the evaluation loss. Using these notations, we aim to design a collaborative inference procedure along with the incorporation of variable number of dedicated trained models, which allow us to successfully recover $\hat{\myY}_{i_t}$ such that at each time instance $t$, the user of index $i_t$ recovers $\hat{\myY}_{i_t}$ based on  the loss measure $\mySet{L}_{e}({\myY}_{i_t} ; \hat{\myY}_{i_t})$.

	\vspace{-0.2cm}
	\subsection{Collaborative Inference Strategies}
	\label{subsec:Inference}
	\vspace{-0.1cm}  	  
	A straight-forward design is to train a single \ac{dnn} with $\Nparameters$ parameters and have each device use it as its local model. 
	Recalling that $\NetMap_{\MyWeights_i}(\cdot)$ is the mapping carried out by the \ac{dnn} model $\MyWeights_i$, the estimate produced is given by $\hat{\myY}_{i_t} =  \NetMap_{\MyWeights_{i_t}}(\myX_{i_t})$. This approach allows each user to infer without requiring any form of connectivity with the remaining devices, and avoids privacy issues as the users do not need to send their data to a centralized server for inference. The drawback of this approach is that while modern edge devices have improved computational capabilities, their compute resources are still limited compared with powerful centralized servers, and thus large networks typically need to be pruned or quantized to be applied on edge devices \cite{cheng2017survey, he2017channel,agustsson2017soft,jacob2018quantization}. As a result, $\Nparameters$ is typically small compared with the amount of parameters used in conventional centralized \acp{dnn}, implying that the inference accuracy may be degraded. 
	
	An alternative strategy which exploits the ability of the users to collaborate during inference is based on partitioning a pre-trained highly-parameterized \ac{dnn} among multiple users. Here, the computation of  $\hat{\myY}_{i_t}$ is divided among all the $\Nusers$ users, such that the resulting estimate corresponds to the output of a highly-parameterized \ac{dnn} with $\Nusers\cdot \Nparameters$ weights. For instance, $\MyWeights_i$ can be set to be the $i$th layer of a \ac{dnn} with $\Nusers$ layers, and thus setting $\hat{\myY}_{i_t} = \big(\NetMap_{\MyWeights_\Nusers} \circ \cdots \circ \NetMap_{\MyWeights_1}\big)(\myX_{i_t})$ implements highly-parameterized \ac{dnn}-based inference of potentially improved performance \cite[Sec. IV]{chen2019deep}. Nonetheless, this strategy, referred to as {\em sequential offloading}, involves increased delay, due to the need to operate in a sequential manner, and requires connectivity among all users. For instance, when the edge devices are \ac{ai}-aided vehicles, this implies that the same group of vehicles must be located in the same physical area whenever one has to carry out inference.  In fact, such forms of collaboration are typically proposed for settings where inference is to be carried out by a central server rather than on the edge device, which only handles part of the processing locally before sending the features, namely, the output of an internal layer of a deep model, to the server (see, e.g., \cite{kang2017neurosurgeon, eshratifar2019jointdnn,chen2019toward}). The main gains in collaboration in such centralized settings, which differ from  inference on the edge as formulated in Subsection~\ref{subsec:Problem}, follow from the fact that conveying processed features as opposed to sending the sample $\myX_{i_t}$ to the server facilitates compression over rate limited links and alleviates some privacy concerns, in addition to balancing the computational load~\cite{bajic2021collaborative}.

	The relatively intuitive approaches discussed above reveal some key guidelines that should be accounted form in tackling the problem formulated in Subsection~\ref{subsec:Problem}:
	\begin{enumerate}[label={\em G\arabic*}]   
		\item \label{itm:ind}  Each model  $\MyWeights_i$ must correspond to a dedicated \ac{dnn} capable of inferring on its own in order to allow operation without any connectivity.
		\item \label{itm:diverse} These individual \acp{dnn} should be diverse, e.g., if  $\MyWeights_i = \MyWeights_j$ for some $i\neq j$ then users $i$ and $j$ have no added value in collaborating. Thus diversity is key to benefiting from collaboration among devices during inference.
		\item \label{itm:comp} Collaboration should be accompanied with inference-oriented protocols based on the exchange of compressed features. In such way we can facilitate collaboration over rate-limited networks compared to exchanging raw samples.
	\end{enumerate}
	Guidelines \ref{itm:ind}-\ref{itm:diverse} motivate utilizing architectures based on deep ensembles \cite{sagi2018ensemble}, briefly reviewed in the following subsection, while \ref{itm:comp} indicates the potential of incorporating layers with quantized features \cite{van2017neural}, discussed in Subsection~\ref{subsec:Quantization}.

	\vspace{-0.2cm}
	\subsection{Deep Ensembles}
	\label{subsec:DeepEnsembles}
	\vspace{-0.1cm}
	Ensemble methods utilize multiple models whose outputs are combined to achieve improved performance \cite{zhou2012ensemble}. Deep ensembles utilize \acp{dnn} as the individual models. Here, during inference, the input sample is processed by each of these \acp{dnn} in parallel, and their outputs are aggregated into a single prediction \cite{sagi2018ensemble}. Various techniques are proposed in the literature for ensemble aggregation, depending on the overall task, including averaging for regression and classifiers with soft outputs, as well as majority vote for hard-decision classifiers \cite{sagi2018ensemble}. Since these aggregation methods can be applied with different numbers of models, deep ensembles are inherently scalable \cite{lakshminarayanan2017simple}, which is desirable for our edge-based collaborative settings. 
	
	Deep ensembles require the individual models to be diverse, namely, while each predictor is designed for the same task, their individual mapping is different. Various measures have been proposed for quantifying such diversity \cite{liu1999ensemble, kuncheva2003measures}, and different learning algorithms have been suggested for training diverse models. The common  approaches for achieving diversity have each model trained with different data, either in parallel (via {\em bagging}) or sequentially (via {\em boosting}) \cite{buhlmann2012bagging}. Yet, one can also train diverse models by formulating a joint regularized objective as in \cite{brazowski2020collective,li2019ensemblenet, shui2018diversity}, or even by simply using different randomized initializations \cite{fort2019deep}. 
	When diversity holds,  ensemble methods are known to achieve improved accuracy over that of the individual models, in a manner which increases with the number  of models \cite{sagi2018ensemble}. 
	
	\vspace{-0.2cm}
	\subsection{Deep Models with Quantized Features}
	\label{subsec:Quantization}
	\vspace{-0.1cm}

	\ac{dnn}-based compression mechanisms have gained increased popularity over the last few years \cite{ma2019image}. These methods compress the input  into a discrete form, mainly for the task of reconstructing it \cite{balle2016end, agustsson2017soft, mentzer2018conditional, mashhadi2020distributed}, but also for the purpose of information extraction \cite{shlezinger2019deep, torfason2018towards}. 
	 These kind of models are usually comprised of trainable encoder which encodes the input data into a continuous latent space followed by a dedicated discretization layer that compresses the features into a discrete latent representation, i.e., a codebook. The discrete codeword is processed by a trainable decoder that performs the desirable task on the compressed output of the encoder. The discretization layer can perform scalar quantization, which is applied element-wise \cite{shlezinger2019deep, balle2016end}, as well as a joint mapping of multiple elements into single discrete representation via vector quantization \cite{agustsson2017soft,van2017neural}.
	
	The main challenge of applying a discretization layer in a \ac{dnn} is the inability to differentiate through continuous-to-discrete mappings, whose gradients are inherently nullified almost anywhere. This limits the ability to simultaneously train  the encoder and the decoder with conventional first order methods, e.g., \ac{sgd} and its variants. Common approaches to tackle this differentiability issue include replacement of the quantization mapping during training with soft  \cite{agustsson2017soft, mentzer2018conditional, shlezinger2020learning} or stochastic  \cite{balle2016end, mashhadi2020distributed} approximations.
	
	{\bf VQ-VAE Model:} 
	A popular architecture for quantizing \ac{dnn} features in a trainable and bit-efficient manner, which is adopted in the sequel, is the \ac{vqvae} model, proposed in  \cite{van2017neural} for discrete representation learning. The \ac{vqvae} architecture is comprised of an encoder-decoder structure with a latent discretization layer implementing vector quantization, and is accompanied with a dedicated training mechanism that allows to jointly train the encoder and the decoder, along with learning of the vector quantization mapping.
	Let us denote the \ac{dnn} parameters of the encoder and the decoder as $\MyWeights_{E}$ and $\MyWeights_{D}$, respectively. For an input data $\myX$, e.g., an image, the continuous output is written as $\myX_{e} = \NetMap_{\MyWeights_{E}}\left(\myX\right)$.
	The VQ-VAE model holds a discrete \ac{cb}, which is an array of vectors of size $d$, where the cardinality of the codebook is $P = |\ac{cb}|$.
	The output of the encoder, $\myX_{e}$, is divided into $m$ vectors of size $d\times 1$, denoted $\myX_{e} = [\myX_e^1,\ldots, \myX_e^m]$. The parameter $d$ determines the number of samples that are jointly quantized, where for $d=1$ the discretization layer is specialized into a trainable scalar quantizer as in \cite{shlezinger2019deep}.
	Then, each vector is mapped into its nearest codeword in \ac{cb} in the $\ell_2$ norm sense, i.e., 
	\begin{equation}
	\myZ^i = \mathop{\arg\min}\limits_{\myVec{e}\in\mySet{Q}} \norm{\myX_{e}^i - \myVec{e}}_2^2.
	\end{equation}
	
	These discrete vectors are stacked to form $\myZ=[\myZ^1,\ldots, \myZ^m]$, and we write  $\myZ = f_{\mySet{Q}}(\myX_e)$. The discrete representation $\myZ$ is	provided as input to the decoder, i.e., $\hat{\myY} = \NetMap_{\MyWeights_{D}}\left(\myZ\right)$. The overall number of bits used for representing the features is thus $B = m\cdot \log_2 P$.
	
\begin{figure}
		\centering
		\includegraphics[width = \figWidth]{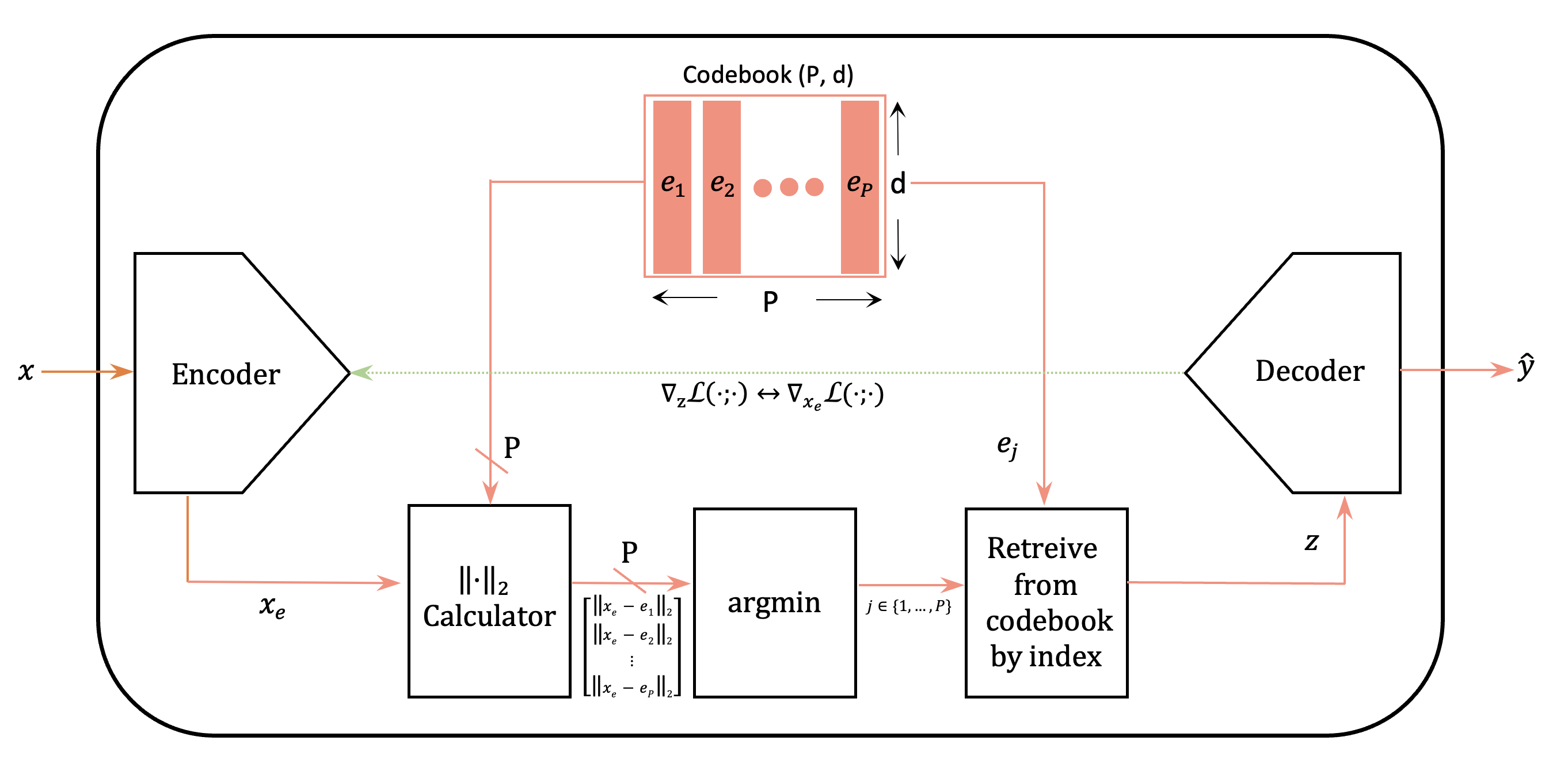}  
		\caption{Vector quantization layer of VQ-VAE.}
		\label{fig:VQVAE}
	\end{figure}

	{\bf VQ-VAE Training:} 
	Training \ac{vqvae} with first-order methods such as \ac{sgd} is impossible since the argmin operation is non-differentiable. In order to tackle this, approximation of the gradient is made by copying it from the decoder input $\myZ$ into the encoder output $\myX_{e}$, i.e., the presence of the codebook is bypassed when computing the gradients via backpropagation. This complete process depicted in Fig.~\ref{fig:VQVAE}, where the codebook bypassing is presented in the dashed line. 

	While bypassing the vector quantizer allows to approximate the gradient of the  training loss, denoted $\mySet{L}({\myY} ; \hat{\myY})$, with respect to $\MyWeights_E$, it does not support the learning of the codebook \ac{cb} in light of the overall task. To tackle this, it was proposed in \cite{van2017neural} to include two additional terms in the  loss function which account for learning \ac{cb}. The first term is the VQ-loss given by $\norm{{\rm sg}\left(\myX_{e}\right)-\myZ}$, which is responsible to move the codebook vectors towards the encoder outputs, where ${\rm sg}$ stands for stop-gradient operator, preventing the gradients from propagating through $\myX_{e}$. The second term is the commitment loss given by $\beta\norm{\myX_{e}-sg\left(\myZ\right)}$, which is intended to bring the encoder outputs closer to the codebook vector. The hyperparameter $\beta$ determines the weight put in for the second term. The overall training loss of the model is written as
	\begin{align}
	\mySet{L}_{tot}({\myY} ; \hat{\myY}) =& \mySet{L}({\myY} ; \hat{\myY}) + \norm{{\rm sg}\left(\myX_{e}\right)-\myZ}_{2}^{2} 
	+ \beta\norm{\myX_{e}-{\rm sg}\left(\myZ\right)}_{2}^{2}.
	\label{eqn:train_loss}
	\end{align}

	Since the codebook size $P$, which is the number of vectors in $\mySet{Q}$, is a design hyperparameter, we can reduce or enlarge the codebook size, and consequently control the number of bits required for representing the codebook vectors indices. The variability of the codebook size and the codebook-vector size is beneficial for keeping the communication efficient when the discrete representation is conveyed over rate-limited channels, as well as preserving the accuracy of an unquantized model, as we consider in the proposed method for collaborative inference over rate-limited channels detailed in the following section.

	
	\vspace{-0.2cm}
	\section{Edge Ensembles}
	\label{sec:EdgeEnsembles}
	\vspace{-0.1cm}
	Here, we present the proposed strategy of edge ensembles for collaborative inference. Unlike offloading-based schemes, our approach is scalable in the number of participating users and allows local inference, while exploiting the presence of collaborative devices to improve accuracy, and supporting low-delay collaboration over rate-limited channels. We begin by presenting the proposed inference mechanism and its underlying rationale in Subsection~\ref{subsec:Rationale}, then we formulate our collaborative inference protocol in Subsection~\ref{subsec:Fusion}, after which we characterize its associated delay and provide a discussion in  Subsections~\ref{subsec:Latency}-\ref{subsec:Discussion}, respectively.

	\vspace{-0.2cm}
	\subsection{High-Level Rationale}
	\label{subsec:Rationale}
	\vspace{-0.1cm} 
	The core idea of edge ensembles is to provide each user the ability to carry out inference on its own, while allowing to benefit from collaboration with neighboring devices by forming together an ensemble of \acp{dnn}. This is achieved by letting the set of local models represent a diverse deep ensemble. Under such a setup, each individual user can utilize its local model for inference, achieving possibly limited accuracy due to the fact that its model is comprised of a compact network with a relatively small number of parameters. Nonetheless, multiple users which are capable of communicating with each other can now collaborate in inference as a form of \ac{ai}-based wisdom of crowds \cite{surowiecki2005wisdom}. In particular, the scalable nature of  ensemble aggregation functions, that can operate with varying number of participating predictors, allows edge ensembles to operate in dynamic and possibly unstable communication conditions, without being dependent on the participation of any specific device in the inference procedure. 
	
	In order to support low-latency inference over rate-limited channels, we adopt \ac{dnn} architectures which incorporate quantized features. We set the models to be divided into encoder-decoder structure with quantized latent features, where the encoder $\MyWeights_E$ is shared among all the devices with its output layer being a trainable quantizer with codebook \ac{cb}. However, the decoders differ between the edge device. The latter guarantees diversity in inference, exploited by the ensemble model, while the usage of a shared encoder with quantized features balances the number of bits the inferring device sends to each neighbor device during inference. The bit rate can be set depending on the achievable rate in the existing links, and to support the required delay in collaboration. 
	

	\vspace{-0.2cm}
	\subsection{Edge Ensemble Inference Protocol}
	\label{subsec:Fusion}
	\vspace{-0.1cm}
	
	To formulate the inference procedure of edge ensembles, let us recall that $\CommLink_{i,j}^{t}$ denotes whether the link between the users $i$ and $j$ at time instance $t$ is active.   
	We define $\mySet{S}_i^t$ as the set of users which user $i$ can reliably communicate with at time $t$, i.e., 	
	\begin{equation}
	\label{eqn:activeNeighb}
	\mySet{S}_i^t \triangleq \{j \in \NusersSet | \CommLink_{i,j}^t \neq 0 \}. 
	\end{equation} 
	By definition $i \in \mySet{S}_i^t$, and $|\mySet{S}_i^t| \in \{1,\ldots,\Nusers\}$. 
	
	The inference stage begins at a given time instance $t$, when each user has a trained deep model available locally. These deep models are comprised of a shared encoder and codebook followed by a user-specific decoder. We denote the shared parameters of the encoder as $\MyWeights_{E}$ and the shared codebook of the quantizer (see Subsection ~\ref{subsec:Quantization}) as $\mySet{Q}$.
	The parameters of the decoders, which are unique and diverse for each edge device, are denoted by $\{\MyWeights_{D,j}\}$.
	As explained in Subsection~\ref{subsec:Problem}, at time instance $t$ the user of index $i_t$ observes a data sample $\myX_{i_t}$, to be used for inference. To assist with its neighbor devices for improving performance, the $i_{t}$th user encodes and quantizes $\myX_{i_t}$ for the purpose of sharing the input while maintaining low latency requirements, resulting in a conveyed quantized version of $\myX_{i_t}$, written as
	\begin{equation}
	\label{eqn:Compfeat}
	\myZ_{i_t}=  \NetMap_{\mySet{Q}}\big(\NetMap_{\MyWeights_{E}}\big(\myX_{i_t}\big)).
	\end{equation}
	The user of index $i_t$ broadcasts the quantized features, $\myZ_{i_t}$, making it available to its active neighbours, i.e., all users in the set $\mySet{S}_{i_t}^t$. Note that when the encoder and quantizer are set to the identity mapping, i.e.,  $\myZ_{i_t} =\myX_{i_t}$, sharing the quantized features boils down to sharing the observed sample  $\myX_{i_t}$. 
	
\begin{figure*}
		\centering
		\includegraphics[width = 1\linewidth]{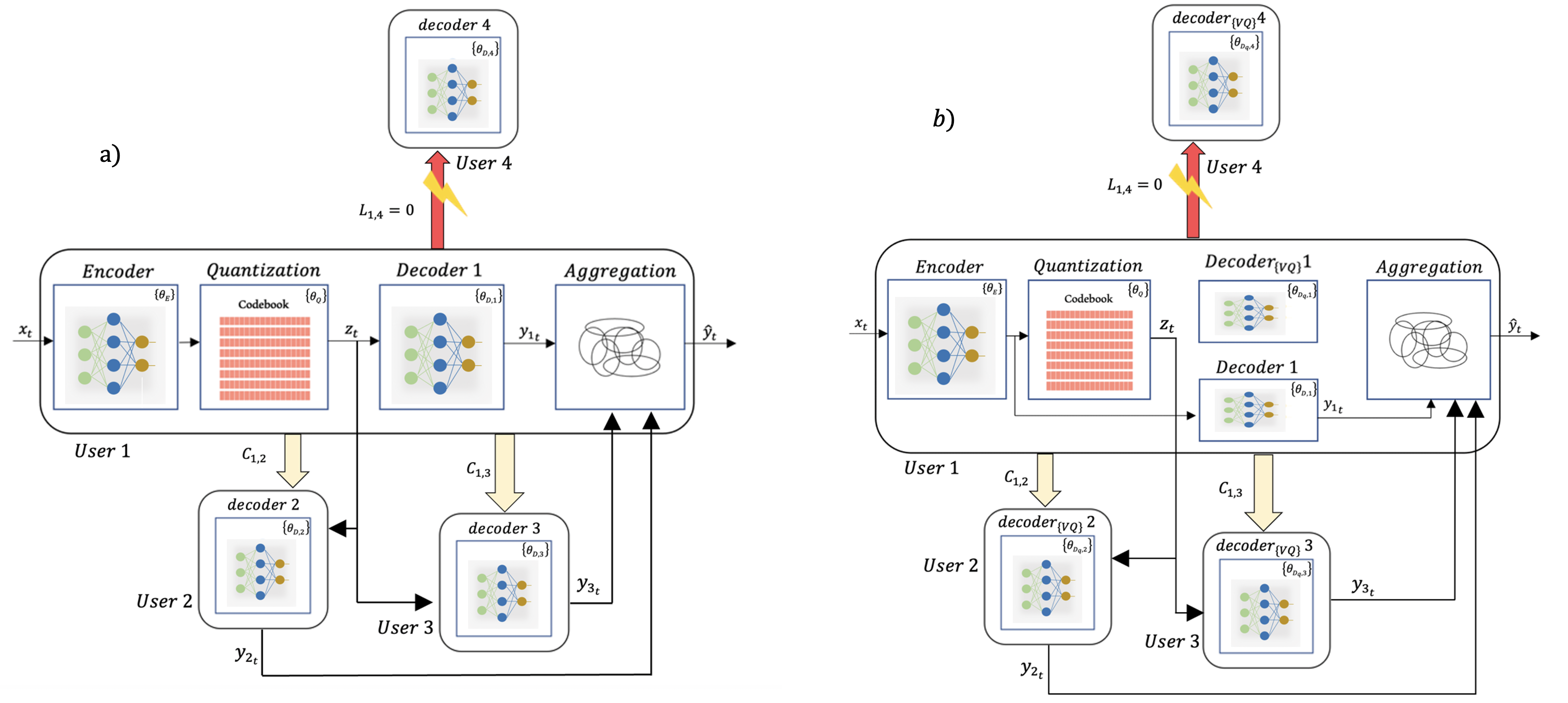}
		\vspace{-0.2cm}
		\caption{Illustration of collaborative inference protocol with shared quantized features, with $K=4$ users, of which one user is not reachable. The quantized features can either be used for inference by all users $(a)$, or solely by the neighbouring devices, while the inferring user processes the input sample $(b)$.}
		\label{fig:SystemModel}
	\end{figure*}

	Next, each user in $\mySet{S}_i^t$ applies its local decoding model to $\myZ_{i_t}$, and conveys the resulting $\NetMap_{\MyWeights_{D,j}}\big(\myZ_{i_t}\big)$ back to user $i_t$, which aggregates them into the predicted  $\hat{\myY}_{i_t}$. 
	In particular, for classification tasks with $\NLabels$ different labels, the output of each individual network $\NetMap_{\MyWeights_{D,j}}\big(\myX_{i_t}\big)$ is an $\NLabels\times 1$ vector whose entries are an estimate of the conditional distribution of each label given $\myZ_{i_t}$, and the predicted $\hat{\myY}_{i_t}$ can be obtained as the label which maximizes the averaged conditional distribution \cite{brazowski2020collective}, i.e.,
	\begin{equation}
	\label{eqn:aggregate}
	\mathop{\arg \max}\limits_{n=1,\ldots,\NLabels} \frac{1}{|\mySet{S}_i^t| } \sum_{j \in \mySet{S}_i^t} \left[ \NetMap_{\MyWeights_{D,j}}\big(\myZ_{i_t}\big)\right]_n. 
	\end{equation}
	 The resulting  inference procedure is summarized as Algorithm~\ref{alg:Algo1} and illustrated in  Fig.~\ref{fig:SystemModel}(a).

	\begin{algorithm}  
		\caption{Collaborative Edge Ensembles Inference with Local Quantized Features}
		\label{alg:Algo1}
		\KwData{Data sample $\myX_{i_t}$ observed at time $t$ at user $i_t$. }
		{ User $i_t$} encodes and quantizes $\myX_{i_t}$ into $\myZ_{i_t}$ via \eqref{eqn:Compfeat}\; 
		User $i_t$ broadcasts $\myZ_{i_t}$ to its neighboring devices\; \label{stp:Comm1}
		\For{each $j \in \mySet{S}_{i_t}^t$  }{
			{User $j$} computes $\NetMap_{\MyWeights_{D,j}}\big(\myZ_{i_t}\big)$\; 
			User $j$ sends the result to user $i_t$\; \label{stp:Comm2}
		}
		{User $i_t$} aggregates $\{\NetMap_{\MyWeights_{D,j}}\big(\myZ_{i_t}\big)\}$ into $\myY_{i_t}$ via, e.g., \eqref{eqn:aggregate}\;
		\KwOut{Prediction $\hat{\myY}_{i_t}$.}
	\end{algorithm}
	
	Algorithm~\ref{alg:Algo1} formulates the collaborative inference rule in which all participating users, i.e., the inferring user $i_t$ and its neighbours, apply their local decoders to the quantized features $\myZ_{i_t}$.  However, while quantizing the features conveyed to the neighbor devices is essential to cope with rate-limited communications, the inferring user is not restricted to produce its local estimate based on the quantized features, and can use the actual sample $\myX_{i_t}$ to which it has access.  In order to support such non-quantized local inference, each user should maintain two different decoders, $\MyWeights_{D,j}$ for inference with the non-quantized version of a locally observed input, and $\MyWeights_{D_q,j}$ for the quantized version of a sample acquired by a neighbour device. In this case, local inference at the $i_t$th user is carried out by applying $\MyWeights_{D,i_t}$ (trained to be applied directly to input samples) to $\myX_{i_t}$, while the neighbouring devices, which have access only to the quantized $\myZ_{i_t}$, use $\MyWeights_{D_q,j}$, as illustrated in Fig.~\ref{fig:SystemModel}(b).

	The non-quantized version of the input, which is available to the inferring user, contains information that may be distorted during quantization. Hence, naive aggregating via, e.g., taking the mean of the estimated probabilities as in~\eqref{eqn:aggregate}, can result in a predication that is less accurate compared with inferring locally. We thus propose a weighted aggregation scheme which balances the significance to the results of the decisions made based on the quantized features by the neighbouring devices, and that made based on the raw sample by the inferring users. To that aim, we replace \eqref{eqn:aggregate} with the following weighted aggregation 
	\begin{equation}
	\label{eqn:aggregate2}
	\mathop{\arg \max}\limits_{n=1,\ldots,\NLabels} \frac{1}{|\mySet{S}_i^t| } \left(\left(\sum_{\substack{j \in \mySet{S}_{i}^{t} \\ {j\neq i}}}\alpha_{j} \left[ \NetMap_{\MyWeights_{D_q,j}}\big(\myZ_{i_t}\big)\right]_n \right)+ \alpha_{i}\left[\NetMap_{\MyWeights_{D,i}}\big(\myX_{i_t}\big)\right]_n\right).
	\end{equation}
	In \eqref{eqn:aggregate2}, the coefficients $\{\alpha_j\}$ are weighting parameters taking values in $[0,1)$, which are set to balance the contribution of the neighbouring devices and that of the local model. 
	These coefficients could be set based on the associated quantization resolution, since the more coarse the quantization is, the lesser the confidence in the decisions based on the quantized features is expected to be. In our experimental study in Section~\ref{sec:Sims} we used the setting of $\alpha_j$ which is based on the number of participating users and the validation accuracy of each user, i.e., users with improved validation accuracy are assigned with higher values of $\alpha_j$, see detailed description in Subsection~\ref{subsec:colab_central3}. The resulting inference procedure, summarized in Algorithm~\ref{alg:Algo2}, allows to assign more confidence to the non-quantized model while still being helped by neighboring devices and benefiting from collaboration. 
	
	\begin{algorithm}  
		\caption{Collaborative Edge Ensembles Inference without Local Quantized Features}
		\label{alg:Algo2}
		\KwData{Data sample $\myX_{i_t}$ observed at time $t$ at user $i_t$; aggregation weights $\{\alpha_{j}\}$}
		{\label{stp:Comm1a} User $i_t$} encodes and quantizes $\myX_{i_t}$ into $\myZ_{i_t}$ via \eqref{eqn:Compfeat}\; User $i_t$ broadcasts $\myZ_{i_t}$ to its neighboring devices\;
		User  $i_t$  computes $\NetMap_{\MyWeights_{D,i_i}}\big(\myX_{i_t}\big)$\;
		\For{each $j \in \mySet{S}_{i_t}^t / i_t$  }{
			{User $j$} computes $\NetMap_{\MyWeights_{D_q,j}}\big(\myZ_{i_t}\big)$\; 
			\label{stp:Comm2a} User $j$ sends the result to user $i_t$\;
		}
		{User $i_t$} aggregates $\{\NetMap_{\MyWeights_{D_q,j}}\big(\myZ_{i_t}\big)\}$ and $\NetMap_{\MyWeights_{D,i_i}}\big(\myX_{i_t}\big)$ into $\myY_{i_t}$ via, e.g., \eqref{eqn:aggregate2}\;
		\KwOut{Prediction $\hat{\myY}_{i_t}$.}
	\end{algorithm}
	

	\vspace{-0.2cm}
	\subsection{Inference Latency Analysis}
	\label{subsec:Latency}
	\vspace{-0.1cm} 	
	Latency is one of the main motivations for utilizing \ac{ai}-empowered edge devices over conventional cloud server-based \acp{dnn}. Our proposed collaborative inference algorithm entails some excessive delay compared with having the users infer only locally, due to the communication overhead and system heterogeneity in the devices. In this section we provide an analysis of the statistical properties of the overall inference latency of Algorithm~\ref{alg:Algo1} (which is similar to that of Algorithm~\ref{alg:Algo2}, as both include the same communications overhead).

	To that aim, we first introduce the following symbols:	
	Let $\{T_{i,j}^{t}\}$ be the communication delays between a pair of distinct users at the beginning of the inference procedure at time $t$. In particular,  $T_{i,j}^{t}$ is the sum of the duration needed for communicating a sample $\myZ_{i}$ from user $i$ to user $j$ at given time $t$, denoted as $\tau_{i,j}$; the computation time required by user $j$ to apply its local decoding model, denoted as $\tau_{P_j}$; and the time required for conveying $\NetMap_{\MyWeights_{D,j}}\big(\myZ_{i}\big)$ back to user $i$ , denoted $\tau_{j,i}$. Therefore, we can write the total delay time in the communication between user $i$ and $j$ as
		\begin{equation}
		\label{eqn:Delay1}
		T_{i,j}^{t} = \tau_{i,j}+\tau_{P_j}+\tau_{j,i}.
		\end{equation}

	The transmitted features vector $\myZ_{i_t}$ is comprised of $\bits_{1}$ bits, which is the number of bits sent by user $i$ to user $j$ at time instance $t$. 	While the channel conditions can vary in time, we consider the conventional model of block-fading temporal variations for wireless channels, where the channels remain fixed during the entire inference stage.
	Hence, the transmitting delay $\tau_{i,j}$ of an active link can be written as $\tau_{i,j}=\frac{\bits_{1}}{\capacity_{i,j}^{t}}$. Similarly, letting $\bits_{2}$ be the number of bits needed to represent the local inference output at device $j$, $\NetMap_{\MyWeights_{D,j}}\big(\myZ_{i_t}\big)$, we can write $\tau_{j,i}=\frac{\bits_{2}}{\capacity_{j,i}^{t}}$. Since the computation delays $\tau_{P,j}$ are dictated by the hardware of each device, which is static, they are assumed here to be deterministic quantities invariant in time. 
	 Consequently, we can write the  delay term \eqref{eqn:Delay1} as
	\begin{equation}
	\label{eqn:Delay2}
	T_{i,j}^{t} = \CommLink_{i,j}^t\left(\frac{\bits_{1}}{\capacity_{i,j}^{t}}+\frac{\bits_{2}}{\capacity_{j,i}^{t}}+\tau_{P,j}\right).  
	\end{equation}
	
	The communication delay in \eqref{eqn:Delay2}  accounts for the possibility of links to be active or non-active, assuming that the inferring user knows with which devices it can communicate. For active links we receive the same term as in \eqref{eqn:Delay1}, while for non-active links there is no latency.
We assume that the channels are reciprocal, such that $\capacity_{i,j}^{t} = \capacity_{j,i}^{t}$, and focus on the inference being a classification task with $\NLabels$ labels for simplicity. In such cases, the response of each neighbouring device, i.e., Step~\ref{stp:Comm2} of Algorithm~\ref{alg:Algo1}, conveys solely an $\NLabels\times 1$ probability vector, which is typically of a much lower dimension compared with the exchanged features $\myZ_{i_t}$ in Step~\ref{stp:Comm1}, and thus $\bits_{1} \gg \bits_{2}$. In such case, the overall delay $T_{i,j}^{t}$ in \eqref{eqn:Delay2} is reduced into 
	\begin{equation}
	\label{eqn:Delay2a}
	T_{i,j}^{t} = \CommLink_{i,j}^t\left(\frac{\bits_{1}}{\capacity_{i,j}^{t}}+\tau_{P,j}\right).  
	\end{equation}
	By using these notations, we can define the overall delay induced by the collaborative inference procedure that starts at time $t$ in Algorithm~\ref{alg:Algo1} as the \ac{rv} $\Delta^{t}$, given by 
	\begin{equation}
	\label{eqn:DeltaDef}
	\Delta^t \triangleq \max\limits_{j\in\NusersSet}\left(T_{i,j}^t\right) = \max\limits_{j\in\NusersSet}\left( \CommLink_{i,j}^t\left(\frac{\bits}{\capacity_{i,j}^{t}}+\tau_{P,j}\right)\right).
	\end{equation}

	The formulation of the inference delay in \eqref{eqn:DeltaDef} allows us to characterize its distribution for i.i.d. links, as stated in the following theorem:
	
	\begin{theorem}
		\label{thm:Latency}
		When the communication links \acp{rv} $\{\CommLink_{i,j}^{t}\}_{i\neq j}$ are i.i.d. with $  \Pr(\CommLink_{i,j}^{t} = 1)\equiv p$, and the channel capacities $\{C_{i,j}^t\}_{i \neq j}$ are positive and i.i.d. \acp{rv} with conditional cumulative distribution function given by $F_C(\epsilon) \triangleq \Pr(C_{i,j}^t < \epsilon |\CommLink_{i,j}^{t} = 1 )$, then for any $\epsilon > 0$, the overall delay \ac{rv} $\Delta^t$ satisfies 
		\begin{equation}
		\label{eqn:Latency}
		\Pr\left(	\Delta^t < \epsilon \right)  = 
		\begin{cases}
		0 & \epsilon \leq \tau_{P_{i_t}}, \\
		\prod_{\substack{j \neq {i}^{t} \\ {\epsilon > \tau_{P_j}}}}\left(1 - p\cdot F_{C}\left(\frac {\bits}{\epsilon-\tau_{P_j}} \right)\right)\prod_{\substack{j \neq {i}^{t} \\ {\epsilon \leq \tau_{P_j}}}}\left(1-p\right) &  \epsilon > \tau_{P_{i_t}}.
		\end{cases} 
		\end{equation}

	\end{theorem}  
	
	\ifFullVersion
	{\em Proof:}
	See Appendix \ref{app:Proof1}. 
	\fi
	
	\smallskip
	Theorem \ref{thm:Latency} reveals some of the statistical properties of the overhead induced by collaborative inference via Algorithm~\ref{alg:Algo1}. For instance, consider the case in which all devices have the same local processing delay, i.e., $\tau_{P,j} \equiv \tau$, and denote the smallest capacity among all the active neighbouring users as $\capacity_{\min} = \min_j\left\{\capacity_{i,j}^t| \CommLink_{i,j}^{t} = 1 \right\}$. Hence, for a fixed number of bits $\bits$ needed to represent $\myZ_{i_t}$, the communication delays are upper bounded by a finite $T_{\max} = \frac{\bits}{\capacity_{\min}}+\tau$. 
	We note that for each $\epsilon \geq T_{\max}$, it holds that $\Pr\left(\Delta^{t} < \epsilon \right)$ equals 1 and $\Pr\left(\Delta^{t} < \epsilon \right) < 1$ for each $\epsilon < T_{max}$. Here, it immediately follows that the delay $\Delta^t$ is not larger than $\frac{\bits}{\capacity_{\min}}+\tau$, which indicates that the resulting delay is smaller than that obtained using edge computing based on sequential offloading with \cite{mao2017survey}. This follows from the fact that the computations of each of the \acp{dnn} in a deep ensemble is carried out in parallel, without requiring any of the users to wait for another device to finish its local computation before it begins. 
  Nonetheless, Theorem~\ref{thm:Latency} also indicates that the delay is likely to increase as the number of users grow, as stated in the following corollary:

	\begin{corollary}
		\label{cor:latency}
		When the computation time is identical among all the devices, i.e., $\tau_{P,j} \equiv \tau$ for each $j \in \NusersSet$, the communication delays are upper bounded by  $T_{\max}$, and $p>0$, then for each  $\epsilon < T_{\max}$ it holds that
		\begin{equation}
		\label{eqn:latency2}
		\lim_{\Nusers\rightarrow\infty} 	\Pr\left(	\Delta^t < \epsilon \right)  = 0 . 
		\end{equation}
	\end{corollary}
	
	\ifFullVersion
	{\em Proof:}
	See Appendix \ref{app:Proof2}.
	\fi
	
	\smallskip
	Corollary \ref{cor:latency} implies that the aforementioned upper bound on the overall latency becomes tight as the number of users grow under identical computation times. Still, this upper bound is typically smaller than the overall delay induced when utilizing sequential offloading, which involves more computation and communication iterations for carrying out inference. Specifically, since the duration of inferring locally, which involves the smallest latency, is $\tau_{P,i}$, Corollary~\ref{cor:latency} suggests that the only additional latency in that case is the maximal one induced by the single communication round in Steps \ref{stp:Comm1} and \ref{stp:Comm2} of Algorithm \ref{alg:Algo1}, i.e., $T_{\max}$.
	
	
	\vspace{-0.2cm}
	\subsection{Discussion}
	\label{subsec:Discussion}
	\vspace{-0.1cm}
	The proposed edge ensembles framework facilitates the operation of deep learning on edge devices by accounting for their unique properties and requirements. In particular, Algorithms \ref{alg:Algo1}-\ref{alg:Algo2} are tailored to account for the limited hardware and computation capabilities of such devices, combined with their mobile dynamic nature and ability to collaborate in a decentralized manner. In fact, edge ensembles treat \ac{ai}-empowered mobile devices as a crowd of diverse intelligent individuals: each agent is capable of inferring on its own, and can thus operate solely without any connectivity. However, a group of users can infer more reliably in a collaborative manner as a form of (artificial) wisdom of crowds \cite{surowiecki2005wisdom}. Specifically, this collaboration boils down to an equivalent deep ensemble, which is an established concept in the deep learning literature \cite{sagi2018ensemble} known to improve accuracy and robustness.   
	
	Collaboration based on edge ensembles is fundamentally different from computation offloading mechanisms, which facilitate complex \ac{dnn}-based computations over a set of edge devices by partitioning a \ac{dnn} and providing each user with a different partition \cite{mao2017survey}. Specifically, computation offloading requires all the users to participate in order to carry out inference, and thus requires constant connectivity, while typically inducing increased latency when the \ac{dnn} partitions must be applied in a sequential manner, e.g., when each partition is a set of layers in a feed-forward network \cite[Sec. IV]{chen2019deep}. Edge ensembles do not rely on connectivity, as each user can infer on its own, and collaboration is applicable with a varying number of users. 
	As Algorithms~\ref{alg:Algo1} and \ref{alg:Algo2} are designed to form a deep ensemble, the more diverse users participate, the more accurate the prediction is expected to be \cite{zhou2012ensemble}. This comes at the probable cost of increased latency, as noted in Subsection~\ref{subsec:Latency}. The accuracy can be thus further improved by allowing more users to participate, e.g., by extending Algorithms \ref{alg:Algo1}-\ref{alg:Algo2} to support multi-hop communications, as done in the context of distributed quantization in \cite{cohen2019serial}, again at the cost of increased latency. 
	
	Collaboration naturally gives rise to challenges that originate from the fact that multiple users participate in inference. These include privacy concerns, due to the exchange of acquired samples; security considerations, as a malicious users can affect prediction; and communication bottlenecks affecting the exchange of information among the participating users. 
	The communication delay of our proposed collaborative inference approach  is reduced by compressing the samples, thus conveying smaller volumes of data. We propose to carry out such compression by implementing vector quantization to the shared samples. The contribution of the quantization step is that the number of bits selected to represent the shared sample is controlled, resulting in overseeing the latency of the communication between the users, which depends on the capacities of the channels. Moreover, quantizing the shared features is likely to contribute to privacy conservation by the fact that the raw samples are not shared, but rather the quantized version of its extracted features \cite{lang2022jopeq}. Security considerations can potentially be alleviated by incorporating Byzantine robust aggregation methods \cite{ghosh2019robust}.
	 Nonetheless,  we leave the analysis of the privacy enhancement capabilities of our approach and its combination with secure aggregation for future work.  While we opt a learned quantization mechanism based on the VQ-VAE architecture in our experimental study in Section~\ref{sec:Sims}, edge ensembles can also be employed with alternative compression schemes for the exchanged features, such as the usage of universal vector quantizers \cite{shlezinger2020uveqfed} or of trainable scalar quantizers \cite{shlezinger2019deep}.

	Designing separate \ac{dnn}-aided edge devices to form a deep ensemble is a consideration that has to be accounted for in the training of the models. For once, the individual \acp{dnn} have to be different from one another in order to benefit from collaboration \cite{kuncheva2003measures}. This can be achieved by either training the \acp{dnn} jointly while boosting diversity, e.g., by using a regularized objective as in \cite{brazowski2020collective,shui2018diversity} or different randomized initialization \cite{fort2019deep}. Alternatively, diverse local models can be trained by modifying distributed learning algorithms to result in different individual networks  \cite{shlezinger2020clustered}. In our numerical study reported in Section~\ref{sec:Sims} we use both a centralized training approach where the local models are trained with the same data and diversity is achieved by different initializations, as well as a distributed-oriented training approach in which each model is trained using shared and user-specific data. Both techniques are shown to yield diverse models allowing the users to benefit from forming an edge ensemble. 
	
	%
	%
	%
	%
	%
	%

	\vspace{-0.2cm}
	\section{Experimental Study}
	\label{sec:Sims}
	\vspace{-0.1cm}
	In this section we numerically evaluate the proposed collaborative inference framework. We begin by evaluating the latency analysis in Subsection~\ref{subsec:latency_eval}, after which we examine the   
	 accuracy of edge ensembles collaborating via Algorithms \ref{alg:Algo1} and \ref{alg:Algo2}. For the latter, we study the gains of collaboration over centralized inference in Subsection~\ref{subsec:colab_central}, and then we evaluate Algorithms \ref{alg:Algo1} and \ref{alg:Algo2} in Subsections~\ref{subsec:colab_central2} and \ref{subsec:colab_central3}, respectively.
	
	For all our experiments we adopt the MobilenetV2 \ac{dnn} architecture \cite{sandler2018mobilenetv2}, being a deep model with multiple layers that is designed to be used on edge devices. We focus on image classification tasks using the CIFAR-10, CIFAR-100, and Imagewoof datasets\footnote{The source code used in our experiments is available at \url{https://github.com/MayMalka10/Ensembles-On-The-Edge}}. We examine the performance with $K \in \{1 \dots 16\}$ devices in which $\{\CommLink_{i,j}^t\}$   are  i.i.d. with connectivity probability $p\in[0,1]$.
	
	\vspace{-0.2cm}
	\subsection{Latency Evaluation}
	\label{subsec:latency_eval}
	\vspace{-0.1cm}
	
	We begin by numerically evaluating the latency analysis derived  in Subsection~\ref{subsec:Latency} in order to visualize the effects of different parameters on the overall latency. Here, the local computation time $\tau_{P,j}$ is   identical among all the devices, and is set to  $700$ milliseconds, being identified in \cite{luo2020comparison} as the MobilenetV2 inference time on various light-weight devices. We evaluate the latency distribution characterized in Theorem~\ref{thm:Latency} when  the capacity \acp{rv} $\capacity_{i,j}^t$ obey an i.i.d.  Rayleigh distribution. The connectivity probability $p$ takes values in $\{0.2,0.8\}$; the number of bits to communicate $B_1$ examined is either $32$ or $128$ bits; and the number of users participating in the collaborative inference is $K=4$ and $K=64$.
	\begin{figure}
		\centering
		\includegraphics[width=0.7\columnwidth]{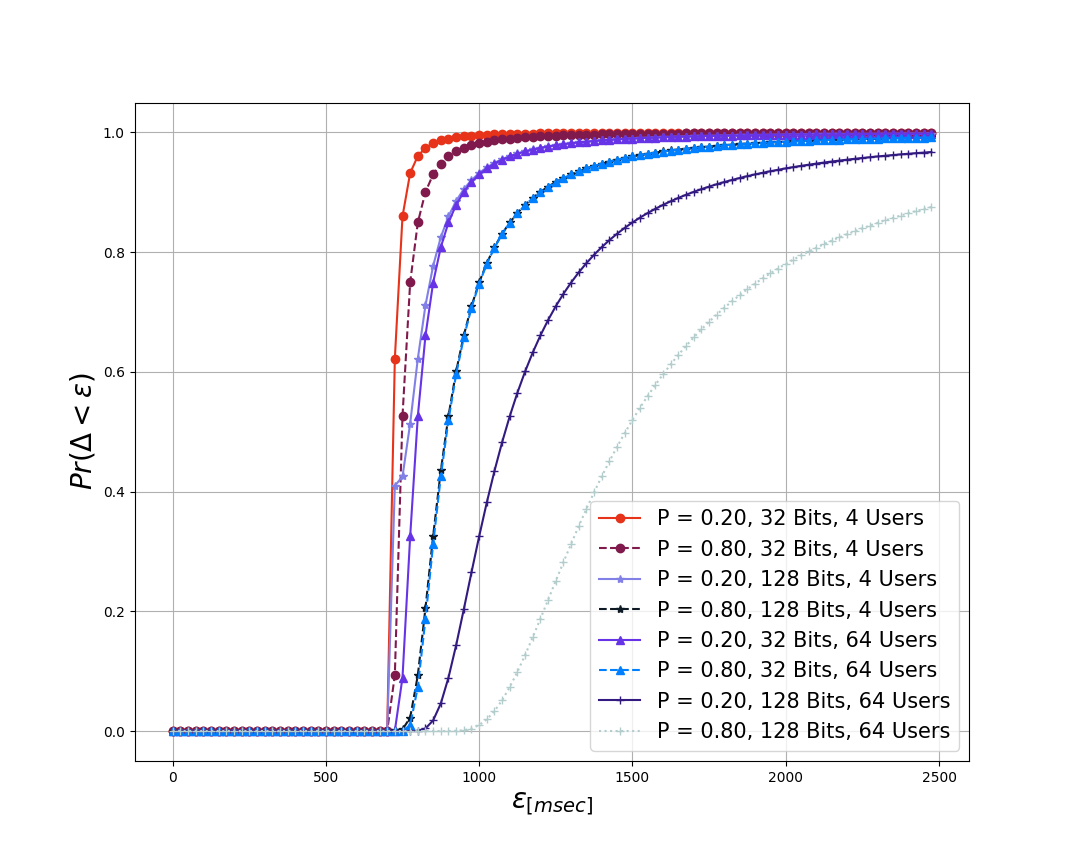}
		\caption{$\Pr(\Delta^t < \epsilon)$ versus the latency in milliseconds.}
		\label{fig:latency_graph}
	\end{figure}
	
	The resulting latency distributions are depicted in Fig.~\ref{fig:latency_graph}. We first observe in Fig.~\ref{fig:latency_graph} that, as expected, the probability that the delay is smaller than $\epsilon$ tends to zero as the required latency is $\epsilon < \tau_{P,j}$, and that the probability starts to increase as $\epsilon$ grows. We can observe the influence of an accretion in the number of bits or in the number of participating users as a more moderate rise in $\Pr(\Delta^t<\epsilon)$. We also see a trade-off behavior in the case of $p=0.2$ between ($128$ bits, $4$ users) and  ($32$ bits, $64$ users). This indicates that in cases of limited connectivity environments, increasing the number of participating users can bridge the gap of lower number of bits.
	
	\vspace{-0.2cm}
	\subsection{Collaborative Inference vs. Centralized Inference}
	\label{subsec:colab_central}
	\vspace{-0.1cm}
	
	We proceed  to examine the benefits in performance of collaboration among edge devices using compact models compared with using a centralized larger model. Since we do not consider here communications and latency considerations, we do not process the input $\myVec{x}_{i_t}$ prior to its sharing with the collaborating devices. Namely, we utilize Algorithm~\ref{alg:Algo1} with an encoder and quantizer mappings set to the identity function.   Our centralized model is a MobilenetV2 \ac{dnn} with width factor one, which has $\Nparameters = 2.23\cdot10^6$ trainable parameters and is trained from random initialization (no pre-training) with CIFAR-10 dataset. For the edge devices, we train a set of diverse MobilenetV2 \acp{dnn} with width factors of $\frac{1}{4}, \frac{1}{3}$, and $\frac{1}{2}$, for which the number of parameters is  $\Nparameters = \{2.51, 3.96,7\}\cdot10^5$, respectively. The edge networks are trained using the same CIFAR-10 dataset and training algorithm, but with different random weight initialization in order to achieve diverse individual networks \cite{fort2019deep}. These individual edge devices then collaborate during inference via Algorithm \ref{alg:Algo1} where the aggregation is computed via \eqref{eqn:aggregate} with $\myVec{z}_{i_t} =\myVec{x}_{i_t}$.
	 
	\begin{figure}
		\centering
		\begin{subfigure}{.5\textwidth}
			\centering
			\includegraphics[width=1\columnwidth]{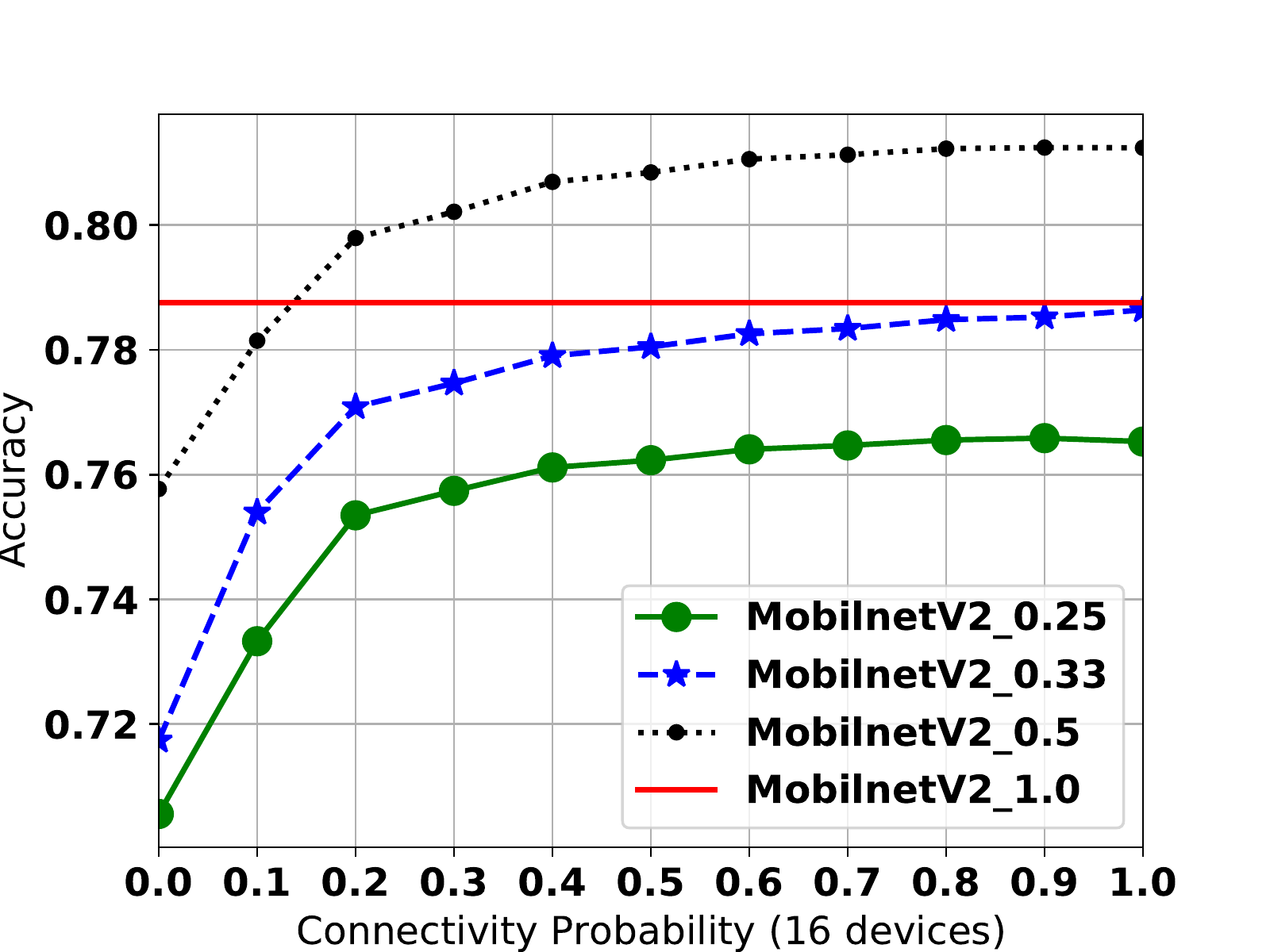}
			\caption{Accuracy versus connectivity probability.}
			\label{fig:n_ensembles_graph}
		\end{subfigure}%
		\begin{subfigure}{.5\textwidth}
			\centering
			\includegraphics[width=1\columnwidth]{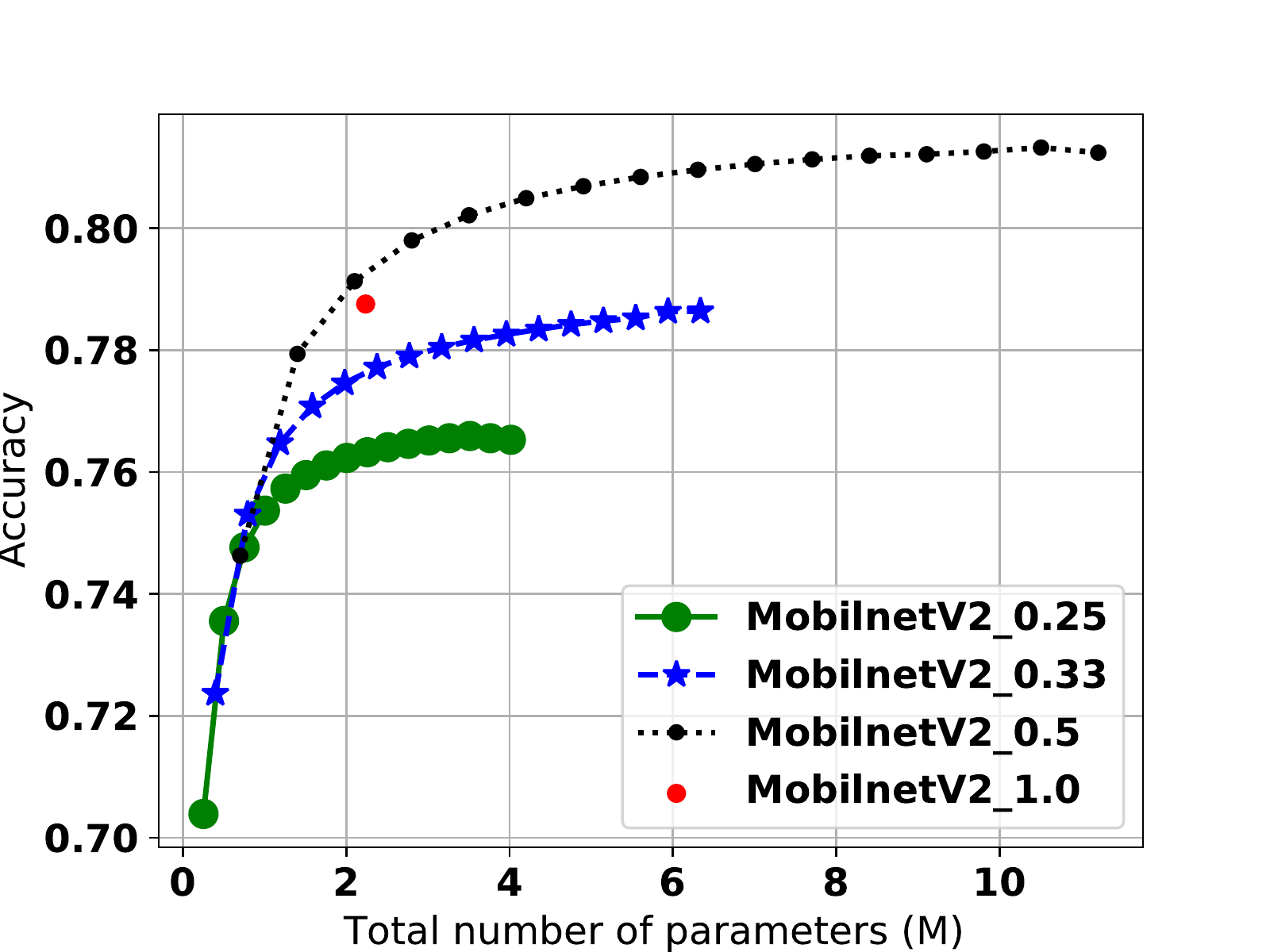}
			\caption{Accuracy versus overall parameters.}
			\label{fig:n_params_graph}
		\end{subfigure}%
		\caption{Accuracy comparison of edge ensembles with different width factors compared with centralized inference with a large network of width factor $1$.}
		\label{fig:ensemble_accuracy}
	\end{figure}
	
	For this setup, we evaluate how the accuracy of edge ensembles grows with the connectivity probability $p$ for a network with $\Nusers=16$ users. The resulting accuracy versus connectivity probability for 16 devices compared to the baseline centralized full MobilnetV2 (abbreviated {\em MobilenetV2\_1.0}) is depicted in Fig. \ref{fig:n_ensembles_graph}. Observing Fig. \ref{fig:n_ensembles_graph}, we note that collaboration among edge devices substantially improves accuracy, while allowing each device to infer solely. For example, having each device utilize a \ac{dnn} with width factor $\frac{1}{2}$, which has roughly $30\%$ the number of weights as MobilenetV2\_1.0, allows each user to achieve inference accuracy of $75\%$ (i.e., for $p=0$), compared to $79\%$ achieved by the full model. However, having multiple devices collaborate via Algorithm \ref{alg:Algo1} improves the accuracy to over $80\%$ when the users can communicate with probability as low as $20\%$. It is also observed in Fig. \ref{fig:n_ensembles_graph} that even when each device has a model as compact as comprising of $6\%$ of the overall number of parameters, which is the case for width factor of $\frac{1}{4}$, the resulting accuracy still improves when $p$ grows, namely, by collaborating via Algorithm \ref{alg:Algo1}, though the collaborating users are unable to outperform the centralized full model, reaching an accuracy of less than $77\%$ for $p = 1$, i.e., full device participation. 
	
	Next, we evaluate how the accuracy scales with respect to the overall number of parameters used for inference, which dictates the capacity of the \ac{dnn}. The results, depicted in Fig. \ref{fig:n_params_graph}, illustrate that collaborative edge ensembles can outperform a centralized model with more parameters than that used by all the participating users together. In particular, we observe that three edge devices equipped with diverse trained  MobilnetV2 \ac{dnn} with width factor $\frac{1}{2}$, whose overall number of weights is $2\cdot 10^6$, outperform the centralized MobilenetV2\_1.0, which has $10^5$ additional parameters. These results indicate that collaborative inference of edge ensembles does not only facilitate inference of \ac{dnn}-based hardware-limited edge devices, but in some scenarios can also allow to achieve improved accuracy with less parameters compared to deep centralized models.

	\vspace{-0.2cm}
	\subsection{Collaborative Inference Protocol via  Algorithm~\ref{alg:Algo1}}
	\label{subsec:colab_central2}
	\vspace{-0.1cm}
	
	In this part, we examine the performance of the proposed collaborative inference protocol when the users communicate over rate-limited channels, and thus share quantized features. We focus here on the case where all users processes quantized features as proposed in Algorithm~\ref{alg:Algo1}. We consider image classification with the CIFAR-100 dataset. The encoder, which is the first five residual blocks of MobilenetV2 (with the standard width factor one), is shared among all participating user alongside the quantizer, that is the trainable codebook of vectors. The decoders are the remaining residual blocks of MobilenetV2, deployed on each edge user, where diversity is achieved in two ways: random initialization and bagging. For the latter we train each \ac{dnn} with different subsets of the training data, where $34000$ out of the $50000$ training samples of CIFAR-100 are available to all users, and $1000$ unique samples are used by each one of the 1$\Nusers=16$ users. This represents settings where training involves small sets of private data available at the users side, as in federated learning, as well as some shared non-private data. The size $d$ of each vector in the quantizer codebook is the number of the encoder output channels, i.e., the encoder output is a tensor of size ${\rm height}\times{\rm width} \times d$, and is divided into ${\rm height}\cdot{\rm width}$ vectors of size $d\times 1$. We test the accuracy achieved by collaborating edge users when quantizing using $\log_2 P \in \{1,4,12\}$ bits, where we achieve different bit settings by dividing each $d\times 1$ sub-vector into up to $8$ parts that are quantized separately.    We  compare the resulting accuracy to that of a single user inferring without quantization. 
	\begin{figure}
		\centering
		\begin{subfigure}{.5\textwidth}
			\centering
			\includegraphics[width=1\columnwidth]{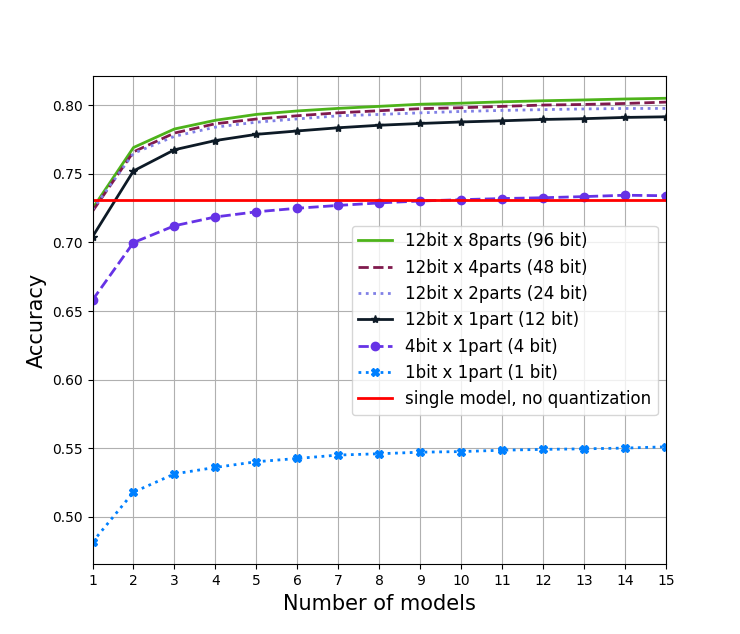}
			\caption{Accuracy versus number of users.}
			\label{fig:cifar100gnAM}
		\end{subfigure}%
		\begin{subfigure}{.5\textwidth}
			\centering
			\includegraphics[width=1\columnwidth]{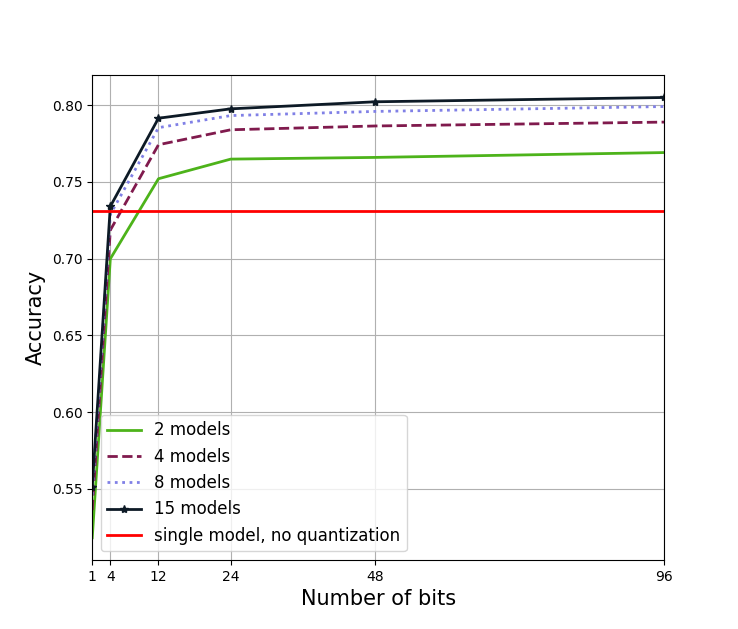}
			\caption{Accuracy versus number of bits.}
			\label{fig:cifar100gnAB}
		\end{subfigure}%
		\caption{Ensemble of edge devices via algorithm \ref{alg:Algo1} - random initialization}
		\label{fig:ensemble_quantized_accuracy_gn}
	\end{figure}
	\begin{figure}
		\centering
		\begin{subfigure}{.5\textwidth}
			\centering
			\includegraphics[width=1\columnwidth]{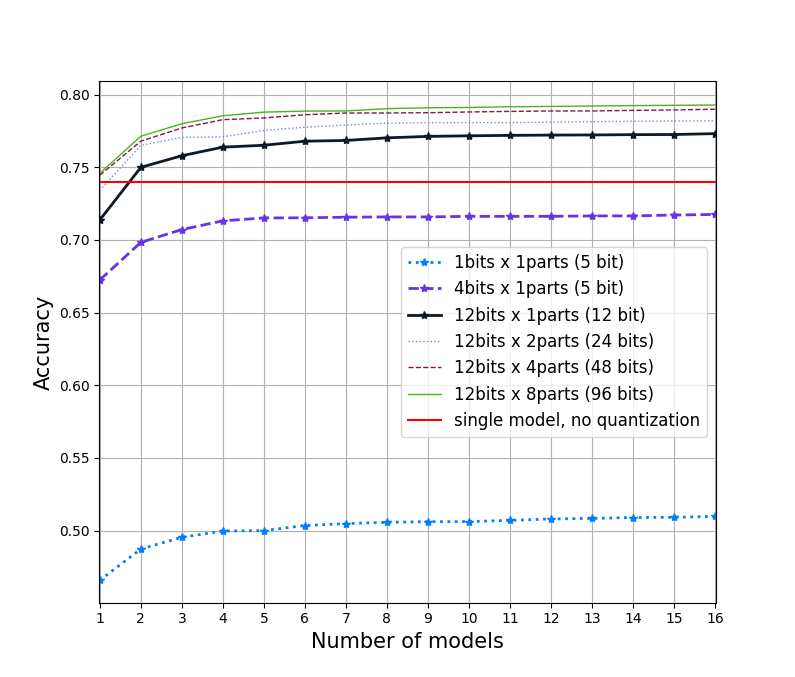}
			\caption{Accuracy versus number of users.}
			\label{fig:cifar100bagAM}
		\end{subfigure}%
		\begin{subfigure}{.5\textwidth}
			\centering
			\includegraphics[width=1\columnwidth]{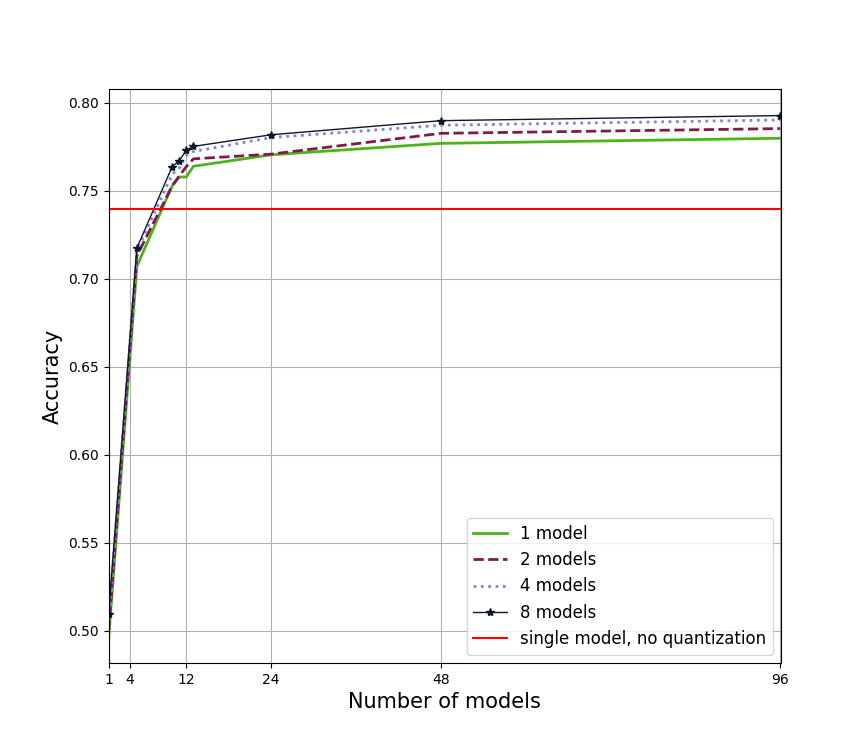}
			\caption{Accuracy versus number of bits.}
			\label{fig:cifar100bagAB}
		\end{subfigure}%
		\caption{Ensemble of edge devices via Algorithm \ref{alg:Algo1} - bagging}
		\label{fig:ensemble_quantized_accuracy_bag}
	\end{figure}

	The accuracy achieved versus number of users and versus number of bits when diversity is achieved via random initialization is depicted in Fig.~\ref{fig:ensemble_quantized_accuracy_gn}.
	From Fig.~\ref{fig:cifar100gnAM} we can see the substantial gains of collaborating with different edge devices by sharing solely quantized features. For instance, a single model with non-quantized features achieves $\sim74\%$ accuracy, while the collaboration of $16$ users with quantized features surpasses an accuracy of $80\%$, namely, a gain of over $6\%$ in accuracy with minor communication overhead.  Moreover, we see that as the number of users grow, the higher the accuracy is, though at a cost of saturated behavior due to limitation in diversity. A similar behavior is observed in Fig.~\ref{fig:ensemble_quantized_accuracy_bag} where diversity is achieved via bagging. In particular, in Fig.~\ref{fig:cifar100gnAB} we observe gains in performance for $\log_2 P \geq 10$ bits when only two users are collaborating, while in Fig.~\ref{fig:cifar100bagAM} we observe that for non-crude quantization, the difference between collaboration among $\Nusers = 10$ users and $\Nusers = 16$ users is minor, which is likely due to the difficulty in yielding sufficiently diverse models using the considered division of data.
	
	\vspace{-0.2cm}
	\subsection{Collaborative Inference Protocol via  Algorithm~\ref{alg:Algo2}}
	\label{subsec:colab_central3}
	\vspace{-0.1cm}
	
	We conclude our experimental study by evaluating Algorithm~\ref{alg:Algo2}, where the aggregating user applies a sepeate decoder \ac{dnn} to the input sample directly. The aggregation accounts for the fact that predictions made by the neighboring users based on quantized features are likely to be less accurate. In particular, we set the aggregation coefficients $\{\alpha_j\}$ in \eqref{eqn:aggregate2} using the validation accuracy values $\{V_j\}$ computed during training. Setting $j=1$ to be the index of the inferring user, $V_1$ is the accuracy achieved with the non-quantized decoder, i.e., $\NetMap_{\MyWeights_{D,1}}\big(\cdot\big)$, while  $V_j$ for $j>1$ is obtained based on the quantized features, namely, with the  mapping $\NetMap_{\MyWeights_{D_q,j}}\big(\NetMap_{\mySet{Q}}\big(\NetMap_{\MyWeights_{E}}\big(\cdot\big)\big)\big)$.
	
\begin{figure}
		\centering
		\begin{subfigure}{.5\textwidth}
			\centering
			\includegraphics[width=1\columnwidth]{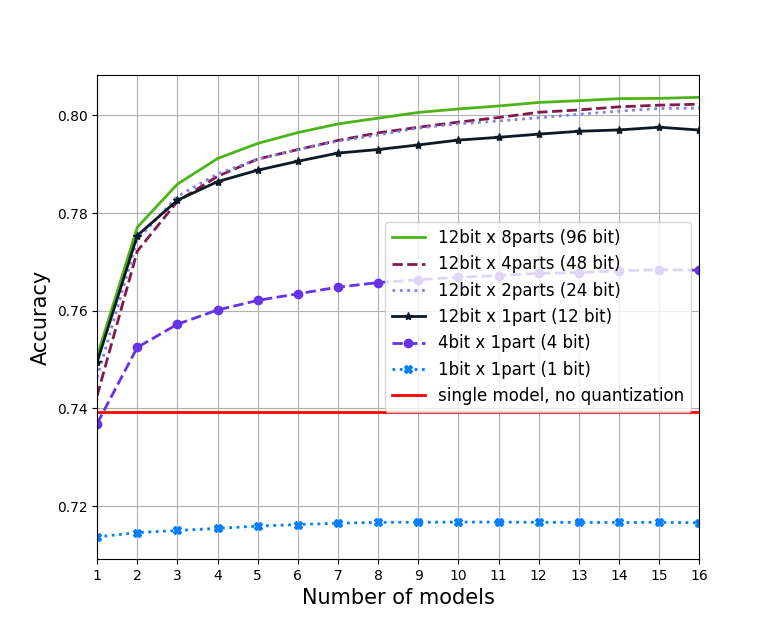}
			\caption{Accuracy versus number of users.}
			\label{fig:cifar100gnAM_algo2}
		\end{subfigure}%
		\begin{subfigure}{.5\textwidth}
			\centering
			\includegraphics[width=1\columnwidth]{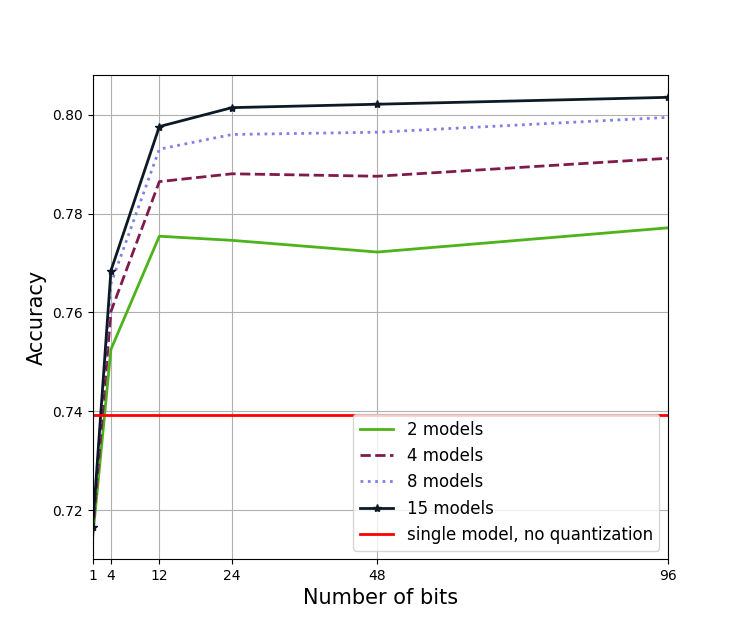}
			\caption{Accuracy versus number of bits.}
			\label{fig:cifar100gnAB_algo2}
		\end{subfigure}%
		\caption{Ensemble of Edge devices via Algorithm \ref{alg:Algo2} - CIFAR100}
		\label{fig:ensemble__accuracy_uq_cifar}
	\end{figure}

\begin{figure}
		\centering
		\begin{subfigure}{.5\textwidth}
			\centering
			\includegraphics[width=1\columnwidth]{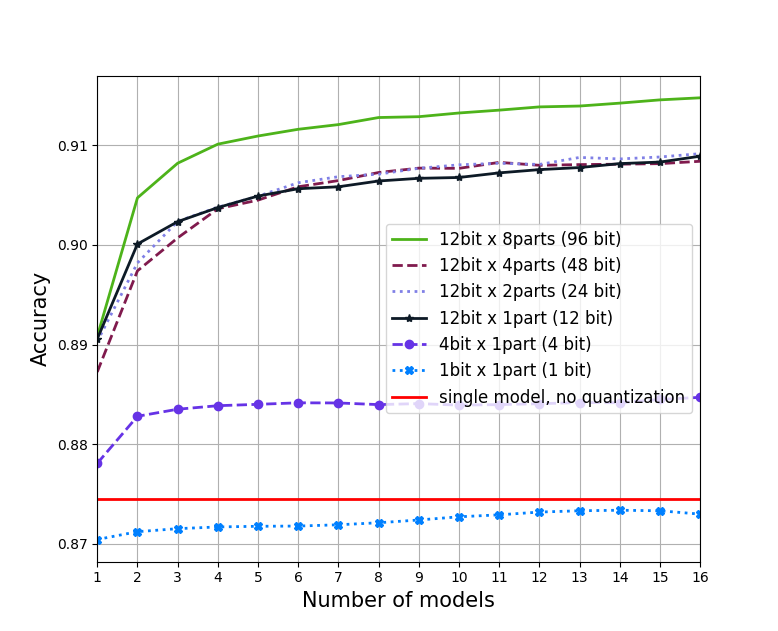}
			\caption{Accuracy versus number of users.}
			\label{fig:WoofGnAM}
		\end{subfigure}%
		\begin{subfigure}{.5\textwidth}
			\centering
			\includegraphics[width=1\columnwidth]{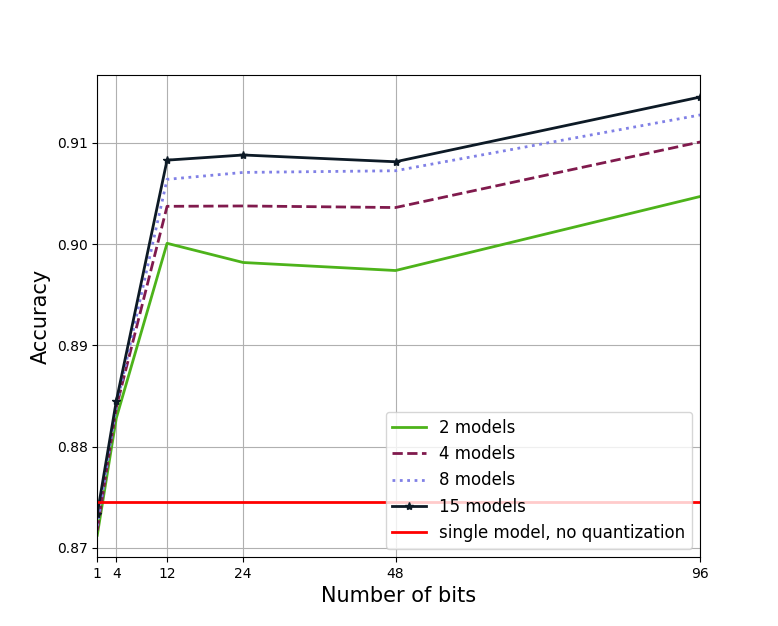}
			\caption{Accuracy versus number of bits.}
			\label{fig:WoofGnAB}
		\end{subfigure}%
		\caption{Ensemble of Edge devices via Algorithm \ref{alg:Algo2} - Imagewoof}
		\label{fig:ensemble__accuracy_uq_woof}
	\end{figure}

	In particular, we set the aggregation coefficients in two steps. First, according to each model validation accuracy, $V_j$, we compute relative accuracy of each device as $\hat{V}_j=\frac{V_j^\rho}{\sum_{i=1}^\Nusers V_i^\rho}$ where $\rho$ is a sharpening constant which we empirically set to $8$. Then, the total aggregation is performed with weighting coefficients
	\begin{equation}
	\label{eqn:AggSim1}
	 \alpha_j = \frac{1}{V_1 + \frac{1}{\sqrt{\Nusers}}\sum_{i=2}^{\Nusers}\hat{V}_i} \cdot \begin{cases}
	 V_1 & j = 1, \\
	 \frac{\hat{V}_j}{\sqrt{\Nusers}} & j > 1.
	 \end{cases}  
	\end{equation}
    Here, all models are trained to be diverse with the same data via random initialization, which was shown in the previous subsection to yield similar diversity to that achieved via data division. We evaluate Algorithm~\ref{alg:Algo2} for image classification using CIFAR-100 dataset as well as Imagewoof, which is a subset of 10 dog breed classes from the Imagenet dataset. The results achived with CIFAR-100 and with Imagewoof are reported in Figs.~\ref{fig:ensemble__accuracy_uq_cifar} and~\ref{fig:ensemble__accuracy_uq_woof}, respectively. 

	
	In Fig.~\ref{fig:cifar100gnAM_algo2} we can observe the same behavior as seen in Fig.~\ref{fig:cifar100gnAM} where an ensemble accuracy of slightly more than $80\%$ is reached. Since the first model processes unquantized features,  Algorithm~\ref{alg:Algo2} achieves accurate predictions while balancing the contribution of decisions made by neighbouring devices based on crudely quantized features. In Fig.~\ref{fig:cifar100gnAM}, 4-bit quantization reaches an accuracy of single unquantized model for more than $\Nusers = 10$ participating users, where in Fig.~\ref{fig:cifar100gnAM_algo2} only two models are needed to outperform single unquantized model. The same behavior is reported in Fig.~\ref{fig:cifar100gnAB_algo2} alongside Fig.~\ref{fig:cifar100gnAB}. 
		These results demonstrate that Algorithm~\ref{alg:Algo2} allows to benefit from collaboration and outperform single model even for crude quantization, such as 4-bit quantization, without harming the performance of the single unquantized model. This is most notable when observing that in Fig.~\ref{fig:cifar100gnAM}, $\Nusers=16$ users collaborating with 4-bit quantization via Algorithm~\ref{alg:Algo1} were able to perform similarly as a single unquantized model, where in Fig.~\ref{fig:cifar100gnAM_algo2}, $\Nusers =16$ users improve by $3\%$ the accuracy of single model. This shows the strength of Algorithm \ref{alg:Algo2} in limited connectivity environments over Algorithm~\ref{alg:Algo1}.
	
	While each sample CIFAR-100 is an image of solely $(32\times32)$ pixels, in Imagewoof each sample is comprised of $(256\times256)$ pixels, and thus compression becomes even more critical to reduce communication overhead and delay. Nonetheless, the high dimensionality of this data set indicates that one can potentially apply crude quantization while preserving the ability to reliably infer via Algorithm~\ref{alg:Algo2}, as we observe in Fig.~\ref{fig:ensemble__accuracy_uq_woof}. In particular, as observed in Fig.~\ref{fig:WoofGnAM}, the accuracy of 1-bit quantization reaches $87.4\%$ accuracy for large number of participating models, which is almost as that of a  single unquantized model which achieves $87.5\%$ accuracy. Noting that with the lower-dimensional CIFAR-100, using 1-bit quantizers lead  to bad predictions comparing with an unquantized single model, we conclude that the benefits of the proposed edge ensembles mechanism in allowing decentralized collaborative inference with low latency grow more notable for realistic high-dimensional data.

	\vspace{-0.2cm}
	\section{Conclusions}
	\label{sec:Conclusions}
	\vspace{-0.1cm}
	In this paper we proposed edge ensembles, which is a \ac{dnn}-based mobile edge collaborative inference scheme that is based on  the joint forming of deep ensembles by multiple users. Our proposed edge ensemble mechanism allows \ac{ai}-empowered edge devices to operate at limited connectivity setups, as well as achieve improved accuracy by collaboration with neighboring devices. We have characterized the inference latency of this strategy, showing that it results in a minor overhead compared to local inference. Our numerical tests demonstrate that such a collaboration allows a group of users equipped with compact \acp{dnn} to achieve improved accuracy by collaboration for a variety of data sets, Our results indicate the potential of the proposed strategy in facilitating accurate \ac{dnn}-aided inference on hardware-limited edge devices. 
	
	\ifFullVersion	
	%
	\vspace{-0.2cm}
	\begin{appendix}
		%
		\numberwithin{proposition}{subsection} 
		\numberwithin{lemma}{subsection} 
		\numberwithin{corollary}{subsection} 
		\numberwithin{remark}{subsection} 
		\numberwithin{equation}{subsection}	
		%
		%
		\vspace{-0.1cm}
		\subsection{Proof of Theorem \ref{thm:Latency}}
		\label{app:Proof1}
		\vspace{-0.1cm}
		Since $\CommLink_{i_t,i_t} = 1$ and $T_{i_t,i_t} = 0$ with probability one, it follows from \eqref{eqn:DeltaDef} that $\Delta^t$ is not smaller than $\tau_{P_{i_t}}$ with probability one, i.e., $\Pr(\Delta^t < \epsilon) = 0$ for $\epsilon \leq \tau_{P_{i_t}}$.
		For $\epsilon > \tau_{P_{i_t}}$ it holds that 
		\begin{align}
		\Pr(\Delta^t < \epsilon) 
		=&\Pr\left( \max\limits_{j\in\NusersSet}\left( \CommLink_{i,j}^t\left(\frac{\bits}{\capacity_{i,j}^t} +\tau_{P_j}\right)\right) < \epsilon\right) \notag \\ 
		\stackrel{(a)}{=}&\Pr\left( \max\limits_{j\neq i_t}\left( \CommLink_{i,j}^t\left(\frac{\bits}{\capacity_{i,j}^t}+\tau_{P_j}\right)\right) < \epsilon\right) \notag \\
		\stackrel{(b)}{=}&\prod_{j \neq i_t}\Pr \left( \CommLink_{i,j}^t\left(\frac{\bits}{\capacity_{i,j}^t}+\tau_{P_j}\right) < \epsilon \right) \notag \\
		\stackrel{(c)}{=}&\prod_{j \neq i_t}(1-p) + p\Pr \left(\frac{\bits}{\capacity_{i,j}^t}+\tau_{P_j} < \epsilon | \CommLink_{i_t,j} =1 \right) \notag \\
		\stackrel{(d)}{=}&\prod_{\substack{j \neq {i}^{t} \\ {\epsilon > \tau_{P_j}}}}\left((1-p) + p\Pr \left(\capacity_{i,j}^t > \frac {\bits}{\epsilon-\tau_{P_j}} | \CommLink_{i_t,j} =1 \right)\right)   \prod_{\substack{j \neq {i}^{t} \\ {\epsilon \leq \tau_{P_j}}}}\left(1-p\right) \notag \\
		\stackrel{(e)}{=}&\prod_{\substack{j \neq {i}^{t} \\ {\epsilon > \tau_{P_j}}}}\left((1-p) + p\left(1-F_{C}\left(\frac {\bits}{\epsilon-\tau_{P_j}} \right)\right)\right)\prod_{\substack{j \neq {i}^{t} \\ {\epsilon \leq \tau_{P_j}}}}\left(1-p\right) \notag \\
		\stackrel{(f)}{=}&\prod_{\substack{j \neq {i}^{t} \\ {\epsilon > \tau_{P_j}}}}\left(1 - pF_{C}\left(\frac {\bits}{\epsilon-\tau_{P_j}} \right)\right)\prod_{\substack{j \neq {i}^{t} \\ {\epsilon \leq \tau_{P_j}}}}\left(1-p\right)
		\label{eqn:Proof1}
		\end{align}
		Here, $(a)$ follows since $\CommLink_{i,j}^t\left(\frac{\bits}{\capacity_{i,j}^t}+\tau_{P,j}\right) = \tau_{P_j} < \epsilon$ with probability one; $(b)$ holds since the \acp{rv} $\CommLink_{i,j}^t\left(\frac{\bits}{\capacity_{i,j}^t}+\tau_{P,j}\right)_{j\neq i_t}$ are mutually independent; and $(c)$ stems from the low of total probability and the fact that $\CommLink_{i_t,j}$ takes the values of zero and one with probabilities $1-p$ and $p$, respectively; In $(d)$ we split the equation into two terms, where for $\epsilon < \tau_{P_j}$ the probability term equals zero; $(e)$ is obtained by substituting $F_C(\epsilon)$ instead of $\Pr(C_{i,j}^t < \epsilon |\CommLink_{i,j}^{t} = 1 )$; and $(f)$ follows since capacity is non-negative, therefore $F_C(\mu) = 0$ for $\mu \leq 0$, and particularly for $\mu = \frac {\bits}{\epsilon-\tau_{P_j}} \leq 0$. 
		Equation \eqref{eqn:Proof1} coincides with \eqref{eqn:Latency}, proving the theorem. 	
		\qed
		
		\vspace{-0.1cm}
		\subsection{Proof of Corollary \ref{cor:latency}}
		\label{app:Proof2}
		\vspace{-0.1cm}
		For $\tau_{P,j} \equiv \tau \leq \epsilon$ , it holds that
		\begin{align}
		\label{eqn:lim}
		\prod_{\substack{j \neq {i}^{t} \\ {\epsilon > \tau}}}\left(1-p F_C\left(\frac{\bits}{\epsilon-\tau_{P,j}}\right) \right)
		\!=\! \left(1-p F_C\left(\frac{\bits}{\epsilon-\tau}\right) \right)^{\Nusers\!-\!1}.
		\end{align} 
		Now, for every fixed $\epsilon$ such that $\epsilon < T_{\max}$ it holds that $F_C(\frac{\bits}{\epsilon\! -\! \tau})$ is strictly smaller than one, and thus for every $p > 0 $ we have that $\left(1-p F_C\left(\frac{\bits}{\epsilon-\tau}\right) \right) < 1$ and thus \eqref{eqn:lim} tends to zero as $\Nusers$ grows arbitrarily large.
		For $\tau_{P,j} \equiv \tau > \epsilon$ , it holds that
		\begin{align}
		\label{eqn:lim2}
		\prod_{\substack{j \neq {i}^{t} \\ {\epsilon > \tau}}}\left(1-p\right) = \left(1-p\right)^{\Nusers\!-\!1}.
		\end{align}
		Since $p > 0$ we get that $\left(1-p\right) < 1$ and thus \eqref{eqn:lim2} also tends to zero as $\Nusers$ grows arbitrarily large. Consequently, the product of \eqref{eqn:lim} and \eqref{eqn:lim2} approaches zero as $K\rightarrow \infty$, proving the corollary. 
		\qed
		
	\end{appendix}
	\fi	
	\bibliographystyle{IEEEtran}
	\bibliography{IEEEabrv,refs}

\end{document}